\documentclass{article}

\usepackage{PRIMEarxiv}
\usepackage[utf8]{inputenc} 
\usepackage[T1]{fontenc}
\usepackage[utf8]{inputenc} 
\usepackage[T1]{fontenc}    
\usepackage{hyperref}       
\usepackage{url}            
\usepackage{booktabs}       
\usepackage{amsfonts}       
\usepackage{nicefrac}       
\usepackage{microtype}      
\usepackage{lipsum}
\usepackage{fancyhdr}       
\usepackage{graphicx}       
\graphicspath{{media/}}     
\usepackage{float}
\usepackage{makecell}
\usepackage{multirow}
\usepackage{amsfonts}
\usepackage{graphicx}
\usepackage{amsmath}
\usepackage{amssymb}
\usepackage{booktabs}
\usepackage{algorithm}
\usepackage{algpseudocode}
\algtext*{EndWhile}
\algtext*{EndIf}
\algtext*{EndFor}
\algtext*{EndProcedure}

\title{Medial Spectral Coordinates for 3D Shape Analysis
}

\begin{document}
\maketitle
\vspace{-2cm}
\begin{center}
    \textbf{\large Morteza Rezanejad$^{123}$, Mohammad Khodadad$^{4}$, Hamidreza Mahyar$^{5}$, Herve Lombaert$^{6}$,\\[0.3cm] Michael Gruninger$^{3}$, Dirk Walther$^{1}$, Kaleem Siddiqi$^{7}$}\\[.5cm]
    \textit{$^{1}$Department of Psychology, University of Toronto, Toronto, Canada}\\
    \textit{$^{2}$Department of Computer Science, University of Toronto, Toronto, Canada}\\
    \textit{$^{3}$Department of Mechanical \& Industrial Engineering, University of Toronto, Toronto, Canada}\\
    \textit{$^{4}$Department of Computer Engineering, Sharif University of Technology, Tehran, Iran}\\
    \textit{$^{5}$W Booth School of Engineering Practice and Technology, Hamilton, Canada}\\
    \textit{$^{6}$Département de génie logiciel et des TI, École de technologie supérieure ÉTS, Montréal, Canada}\\
    \textit{$^{7}$School of Computer Science and Centre for Intelligent Machines, McGill University, Montréal, Canada}\\[1cm]
\end{center}

\begin{abstract}
In recent years there has been a resurgence of interest in our community in the shape analysis of 3D objects represented by surface meshes, their voxelized interiors, or surface point clouds. In part, this interest has been stimulated by the increased availability of RGBD cameras, and by applications of computer vision to autonomous driving, medical imaging, and robotics. In these settings, spectral coordinates have shown promise for shape representation due to their ability to incorporate both local and global shape properties in a manner that is qualitatively invariant to isometric transformations. Yet, surprisingly, such coordinates have thus far typically considered only local surface positional or derivative information. In the present article, we propose to equip spectral coordinates with medial (object width) information, so as to enrich them. The key idea is to couple surface points that share a medial ball, via the weights of the adjacency matrix. We develop a spectral feature using this idea, and the algorithms to compute it. The incorporation of object width and medial coupling has direct benefits, as illustrated by our experiments on object classification, object part segmentation, and surface point correspondence.
\end{abstract}

\keywords{Medial Spectral Coordinates, Medial Surface, 3D Shape Analysis}

\section{Introduction}

Advances in manufacturing sensors that can perceive 3D information, including RGBD cameras, are beginning to revolutionize 3D perception using computer vision methods. In this context, 3D object shape analysis has received much attention due to its relevance to a variety of applications in robotics, medical imaging, autonomous driving, surveillance, and other domains. The technologies that are being developed for these applications must grapple with fundamental problems including 3D shape classification, part segmentation, and shape matching. 

The problem of finding correspondences between 3D objects has a long history \cite{biasotti2006sub,brincker2014local,shen1999affine,hilaga2001topology,jain2006robust,lombaert2012focusr,lipman2009mobius}.
Moving beyond object matching, the segmentation of a 3D object into its constituent parts is another classic computer vision problem for which research is very much still active \cite{kalogerakis20173d,lyu2020learning,xu2020learning,thomas2019kpconv,hegde2021pig,rezanejad2015view}. In addition to part segmentation, 3D shape classification remains a topic of active research. On this front, deep learning methods, including
PointNet \cite{qi2017pointnet}, PointNet++ \cite{qi2017pointnet++},
DeepSet \cite{zaheer2017deep}, ShapeContextNet \cite{xie2018attentional}, PointGrid \cite{le2018pointgrid}, DynamicGCN \cite{ye2020dynamic}, and SampleNet \cite{lang2020samplenet}, have shown great promise. 
As such models grow and undergo further development, they also become more data-hungry and require more resources in terms of computing time. Deep neural network models are typically limited by their design and they cannot easily learn features that are not describable by the parts from which they are made. In the context of 
3D shape analysis, it is also important for a neural network based model to be able to learn appropriate latent features when the point cloud is sparse or has been geometrically transformed \cite{xu2020geometry}. 

\begin{figure}
    \centering
    \begin{tabular}{ccccc}
    \includegraphics[height = 0.205\textwidth]{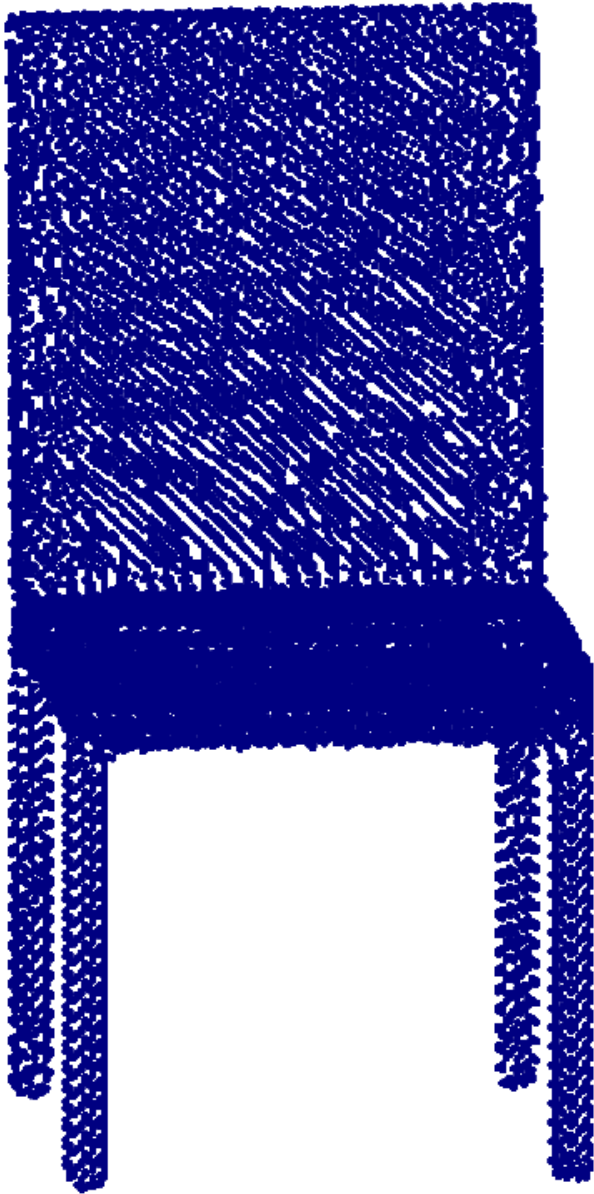}& 
    \includegraphics[height = 0.205\textwidth]{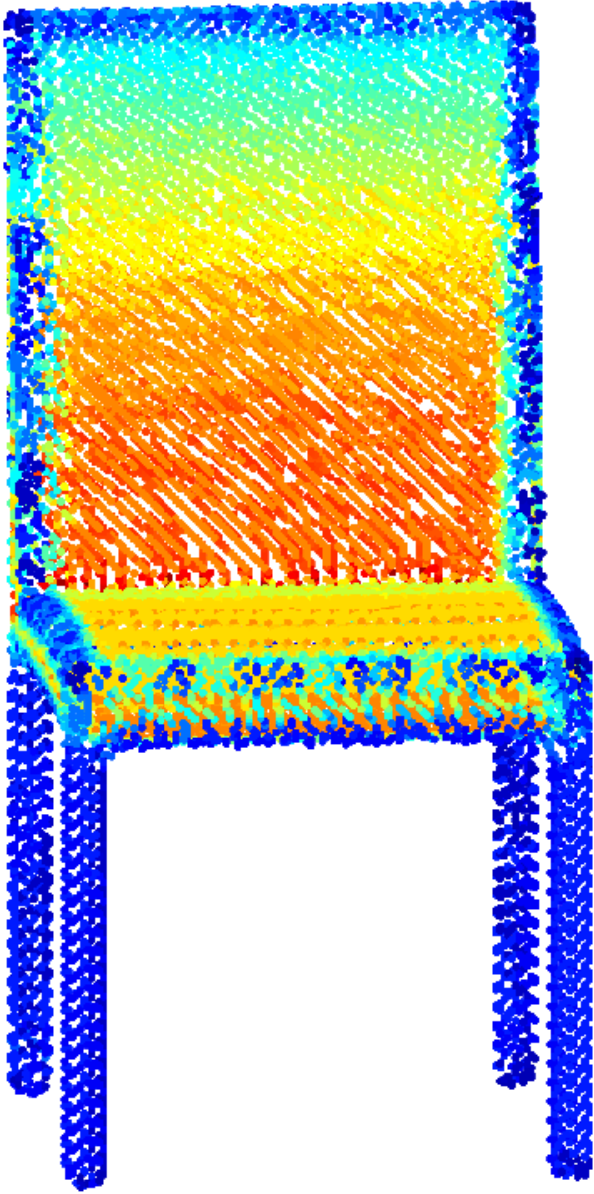} & 
    \includegraphics[height = 0.205\textwidth]{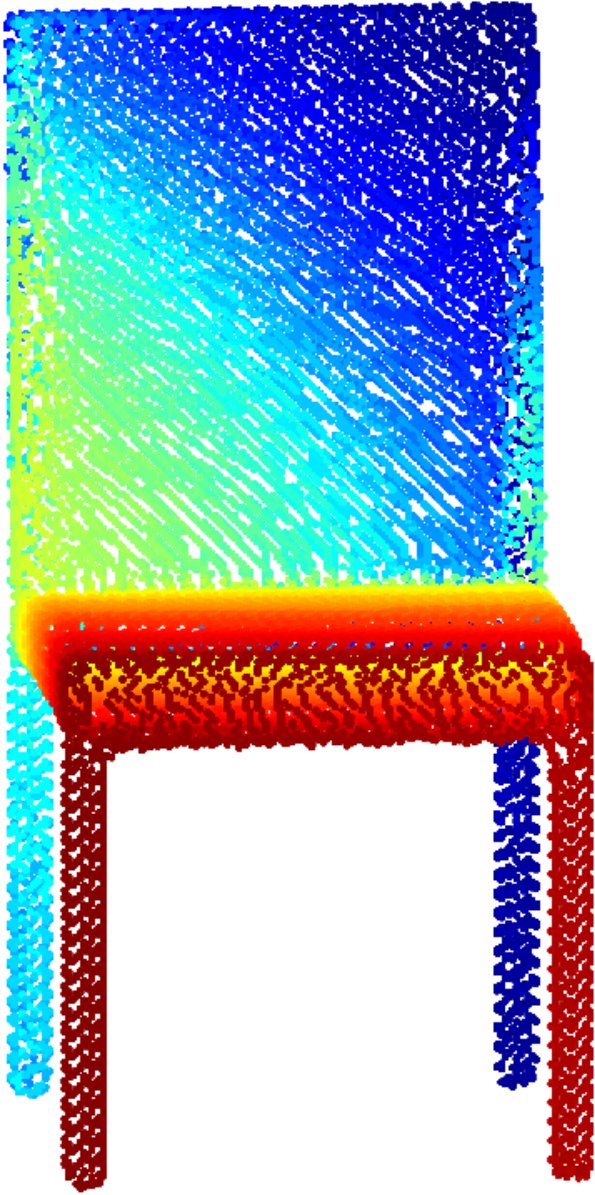} &
    \includegraphics[height = 0.20\textwidth]{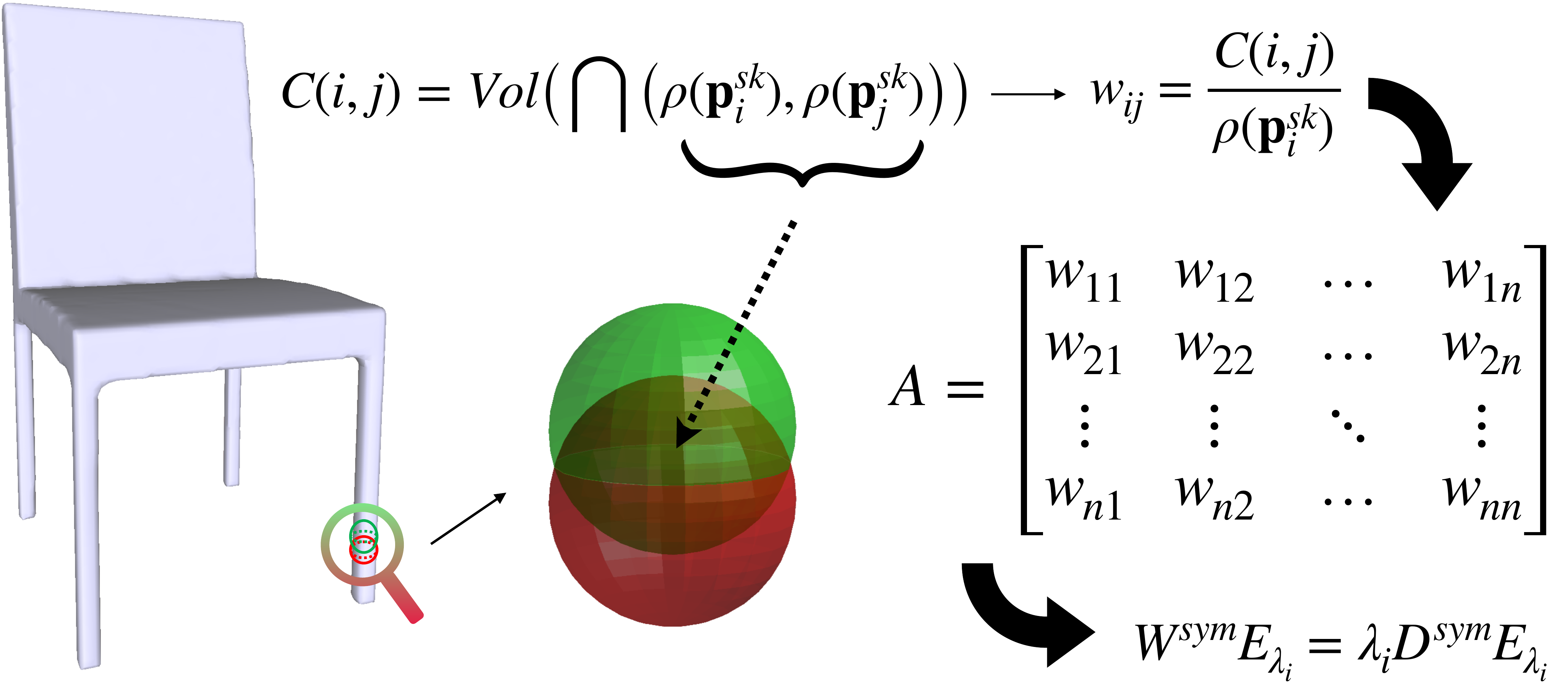} & 
    \includegraphics[height = 0.205\textwidth]{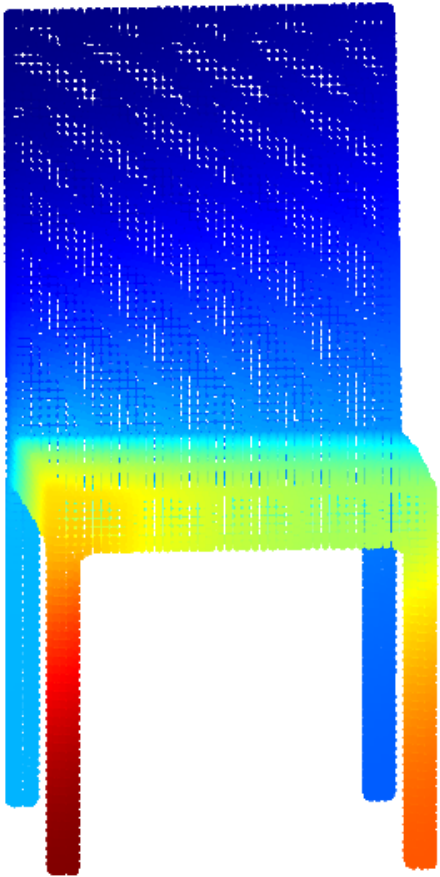}
    \\
    \textbf{(a)} & \textbf{(b)} & \textbf{(c)} & \textbf{(d)} & \textbf{(e)}\\
    \end{tabular}
    \begin{tabular}{ccc}
    \includegraphics[height = 0.215\textwidth]{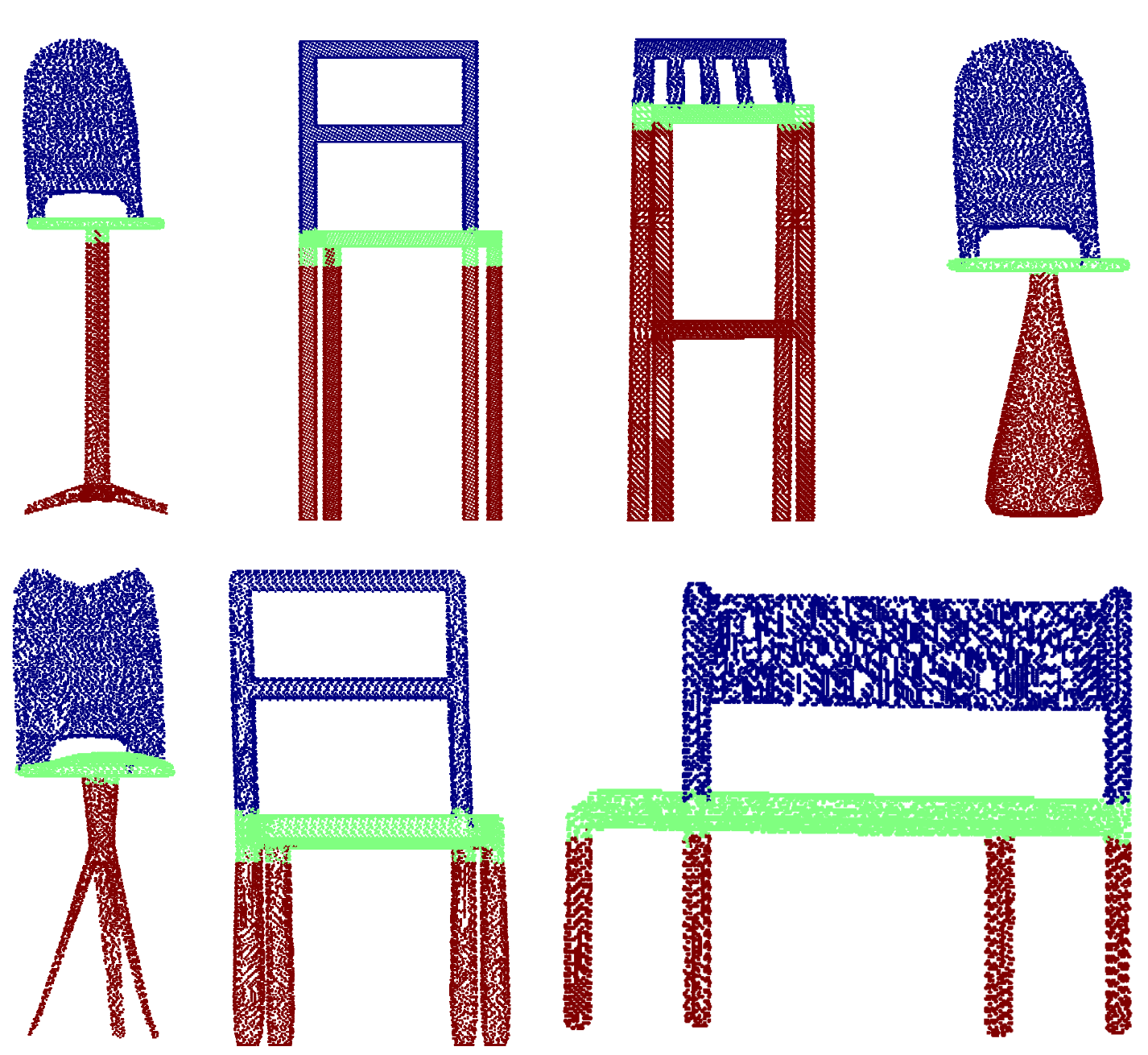}&
    \includegraphics[height = 0.215\textwidth]{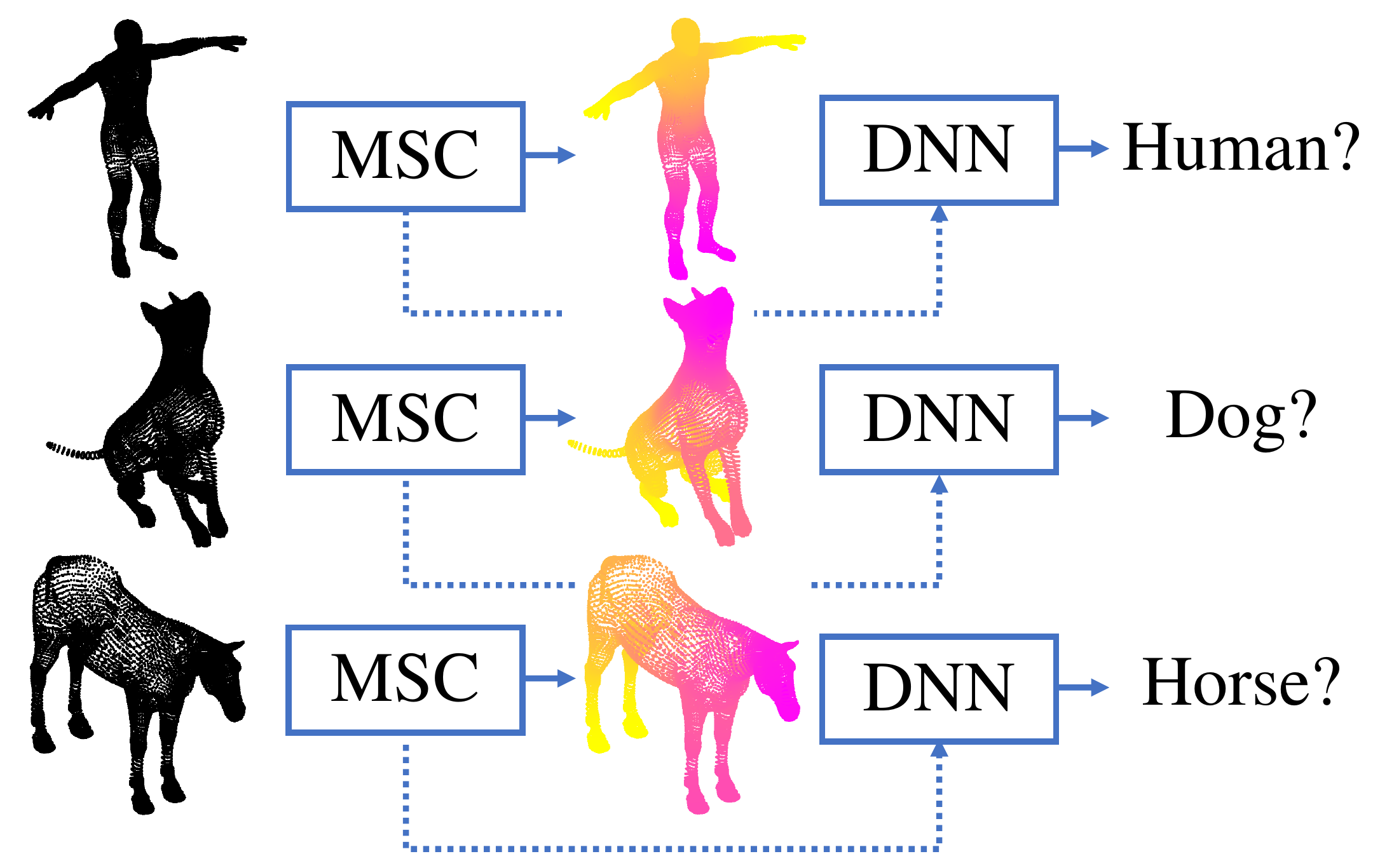}
    &\includegraphics[height = 0.215\textwidth]{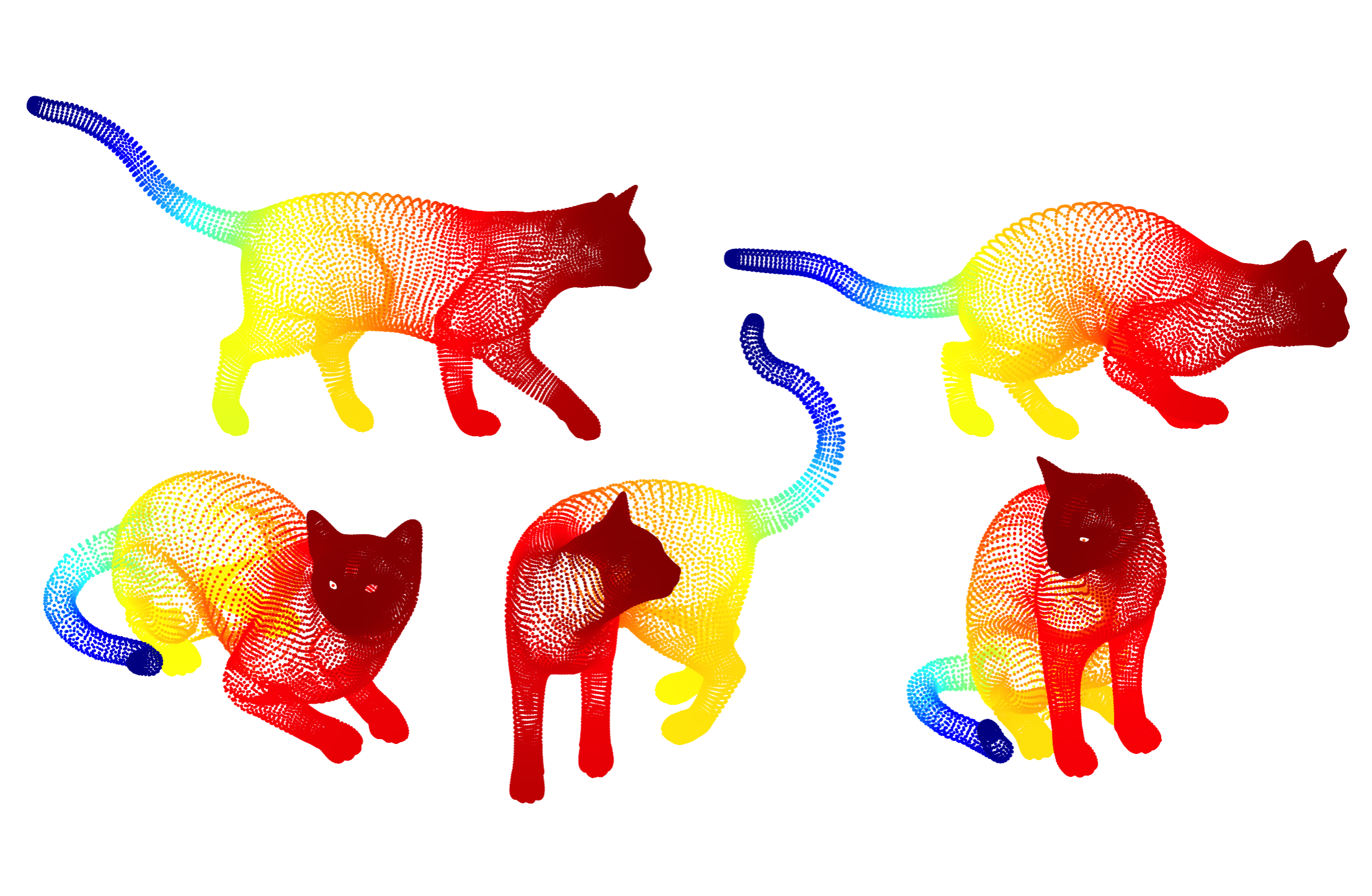}
    \\
    \textbf{(f)} & \textbf{(g)} & \textbf{(h)} \\
    \end{tabular}
    \caption{An overview of our main results. \textbf{First row:} \textbf{(a)} We begin with a 3D object and compute its medial surface. \textbf{(b)} The radius associated with the medial surface is mapped to its boundary.  \textbf{(c)} Boundary points that are associated with the same medial surface point are coupled and shown with the same color. \textbf{(d)} Entries of the adjacency matrix are computed using the intersection volume of the associated medial spheres to drive adjacency weights, for subsequent spectral analysis. \textbf{(e)} The first non-trivial eigenvector is mapped onto the original shape. \textbf{Second row:} The use of our medially driven spectral coordinates for segmentation \textbf{(f)}, object classification \textbf{(g)}, and shape correspondence \textbf{(h)}.}
\end{figure}%

To facilitate these tasks, a descriptive and versatile representation for 3D object shape is key. To this end, in the present article we propose a novel 3D object shape representation method that computes spectral coordinates using the associated 3D medial surface. Medial representations have been used for 2D shape analysis in the literature for decades \cite{xie2008shape, tam2003shape, bai2008skeleton,macrini2011bone,siddiqi1999shock,rezanejad2020medial,rezanejad2015robust}. Applications of medial representations also exist in 3D \cite{siddiqi2008medial,Tagliasacchi2016,hong2017subject}, but challenges exist around computing them reliably and robustly. In addition, the computer vision community has shifted away from the use of hand-crafted features to embrace the use of deep learning based descriptors, learned from geometric data. Taken together, these methods have allowed for better performance in general, in most tasks that the models have been curated for. 

In the present article, we aim to equip present models with an enriched representation of 3D object shape, one that leverages the local width of the object to construct a spectral feature at each of its boundary points. We make three main contributions.
\begin{enumerate}
    \item We develop a robust, and reliable implementation of an algorithm for computing 3D medial surfaces based on an average outward flux measure. 
    \item We introduce a novel approach to computing spectral coordinates, exploiting the duality between the medial surface and the boundary it represents. This approach brings inferred local and global geometrical information to the boundary and allows for an enriched spectral representation of that boundary. In particular, surface points that are bi-tangent to medial spheres that have a common volume are coupled to capture local object part symmetry.
    \item We demonstrate that by using such spectral coordinates that are associated with the medial surface, there are performance benefits for popular 3D shape analysis tasks, including finding correspondences between surface points, 3D object part segmentation, and 3D object shape classification. 
\end{enumerate}

Our article is organized as follows. In Section \ref{sec:medialaxis}, we review average outward medial surfaces and their computation. 
In Section \ref{sec:spectral_coordinate}, we introduce our novel medial spectral coordinates. In Section \ref{sec:applications} we develop a number of applications that take advantage of our these spectral coordinates. Finally, we conclude with a summary of contributions and a discussion of possible future research directions, in Section \ref{sec:conclusion}.

\section{Medial Representation of 3D Shapes}
\label{sec:medialaxis}

\begin{figure}
	\centering
	\includegraphics[width=0.48\textwidth]{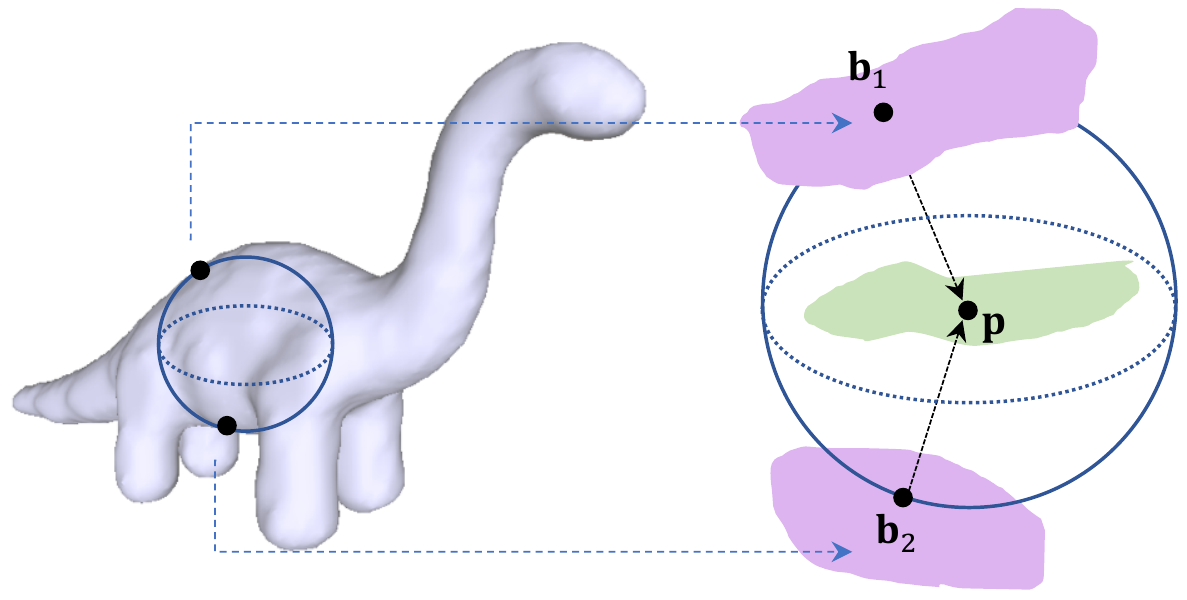}
	\caption{A schematic of a portion of a medial surface together with a selected maximal inscribed sphere, and the associated bi-tangent points ($\mathbf{b}_1$, $\mathbf{b}_2$) where the sphere touches the boundary. The medial surface patch is shown in light green and the associated local boundary patches are in light purple. Each point $\mathbf{p}$ on the medial manifold is associated with two distinct bi-tangent points on the object's surface, to which it is closest in the sense of Euclidean distance (the figure is adapted from \cite{bouix2005hippocampal}).}
	\label{fig:sphere}
\end{figure}


Introduced by Harry Blum \cite{blum1967transformation}, medial axes or skeletons have been used in the vision and computational geometry literature for years \cite{siddiqi2008medial, feldman2006bayesian, siddiqi1999shock, lee1982medial, amenta2001power, tsogkas2017amat, camaro2020appearance,rezanejad2019gestalt}. Formally, the medial axis is defined as the locus of centers of maximal inscribed spheres within a watertight object. In 3D
this locus can include both 3D curves and 3D medial manifolds, that together comprise the 3D medial surface of an object. A promise of this representation is that it captures the duality between both boundary and interior object properties, the latter by considering the radius associated with the inscribed sphere to capture local object width. In addition, it is possible to reconstruct the original object from its medial surface, so in this sense the representation is complete. Unfortunately, in practice, computing a medial surface robustly turns out to be a challenging computational problem. We focus here on the medial surface computation approach of \cite{siddiqi2002hamilton}, which is based on a notion of average outward flux (AOF), and which was later used for hippocampal shape analysis \cite{bouix2005hippocampal} as well as other applications \cite{siddiqi2008medial}.


\subsection{Average Outward Flux Medial Surfaces}
 
\begin{figure*}[!ht]
\centering
\begin{tabular}{c@{\hskip 10pt}c@{\hskip 10pt}c@{\hskip 10pt}c}
    \includegraphics[height=0.18\textwidth,width = 0.21\textwidth]{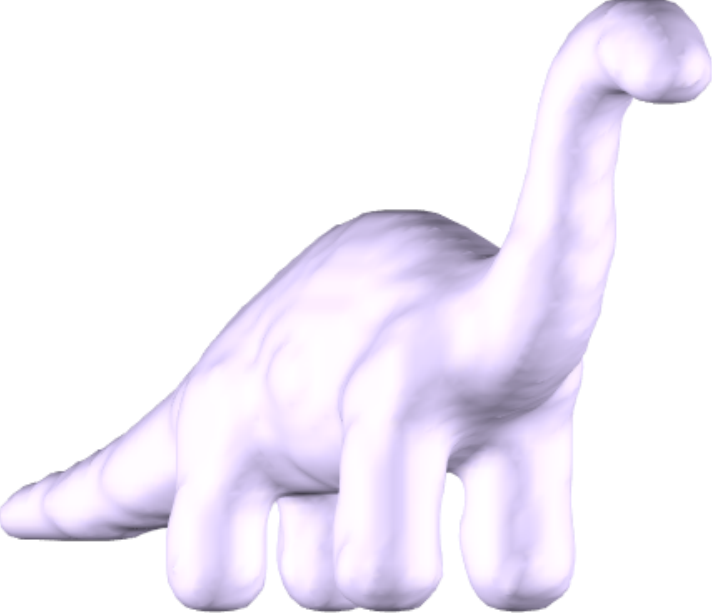} & \includegraphics[height=0.18\textwidth,width = 0.21\textwidth]{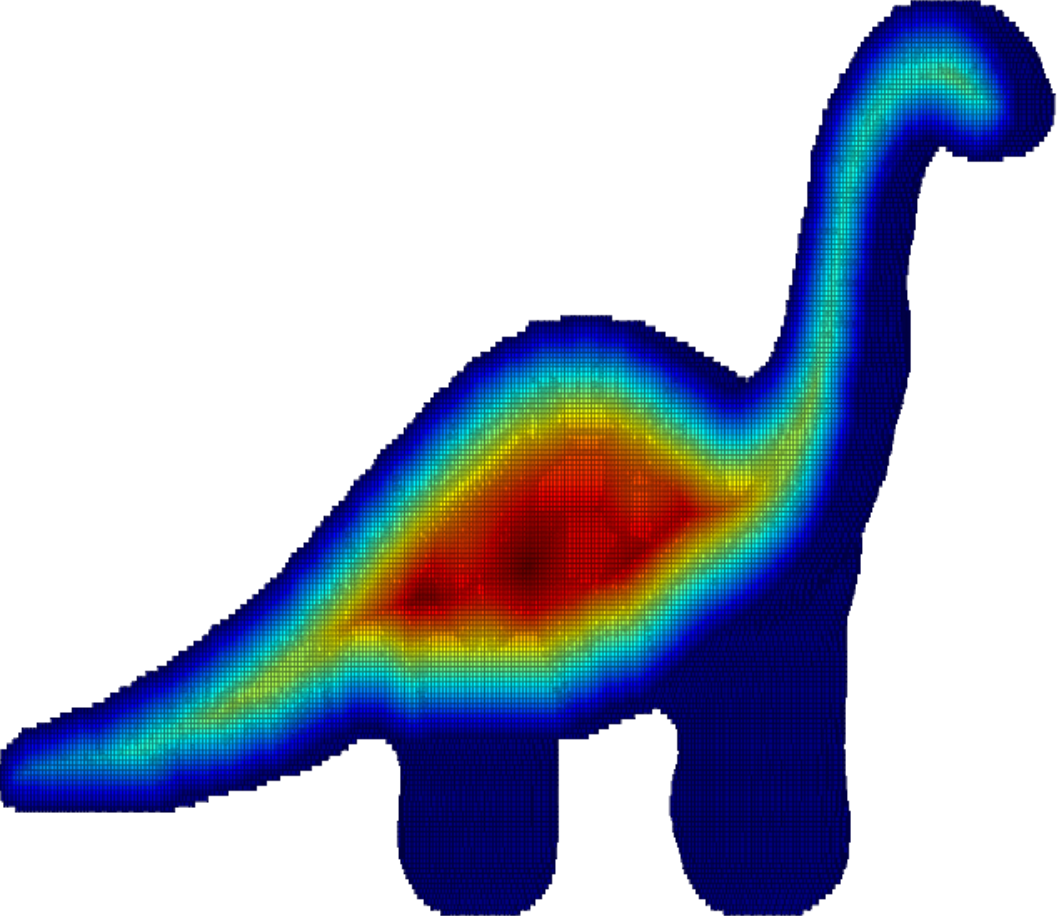} & \includegraphics[height=0.18\textwidth,width = 0.21\textwidth]{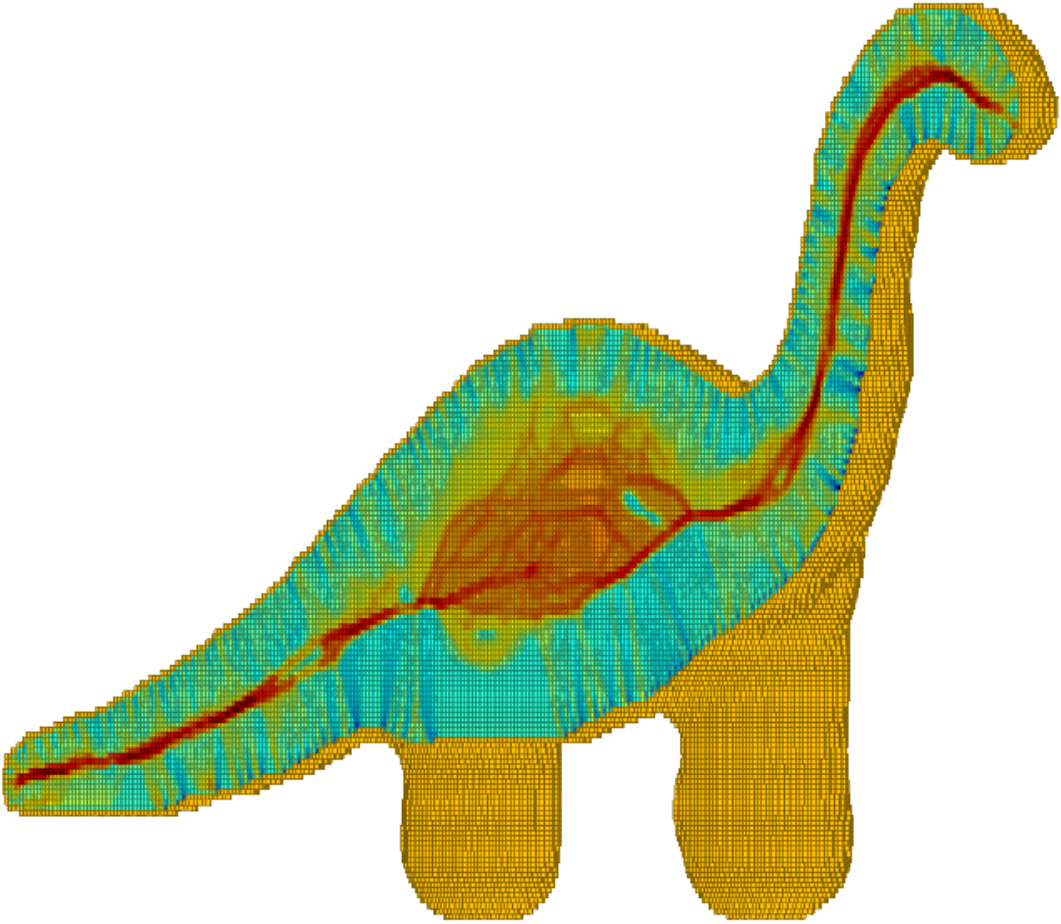} &  \includegraphics[height=0.175\textwidth,width = 0.21\textwidth]{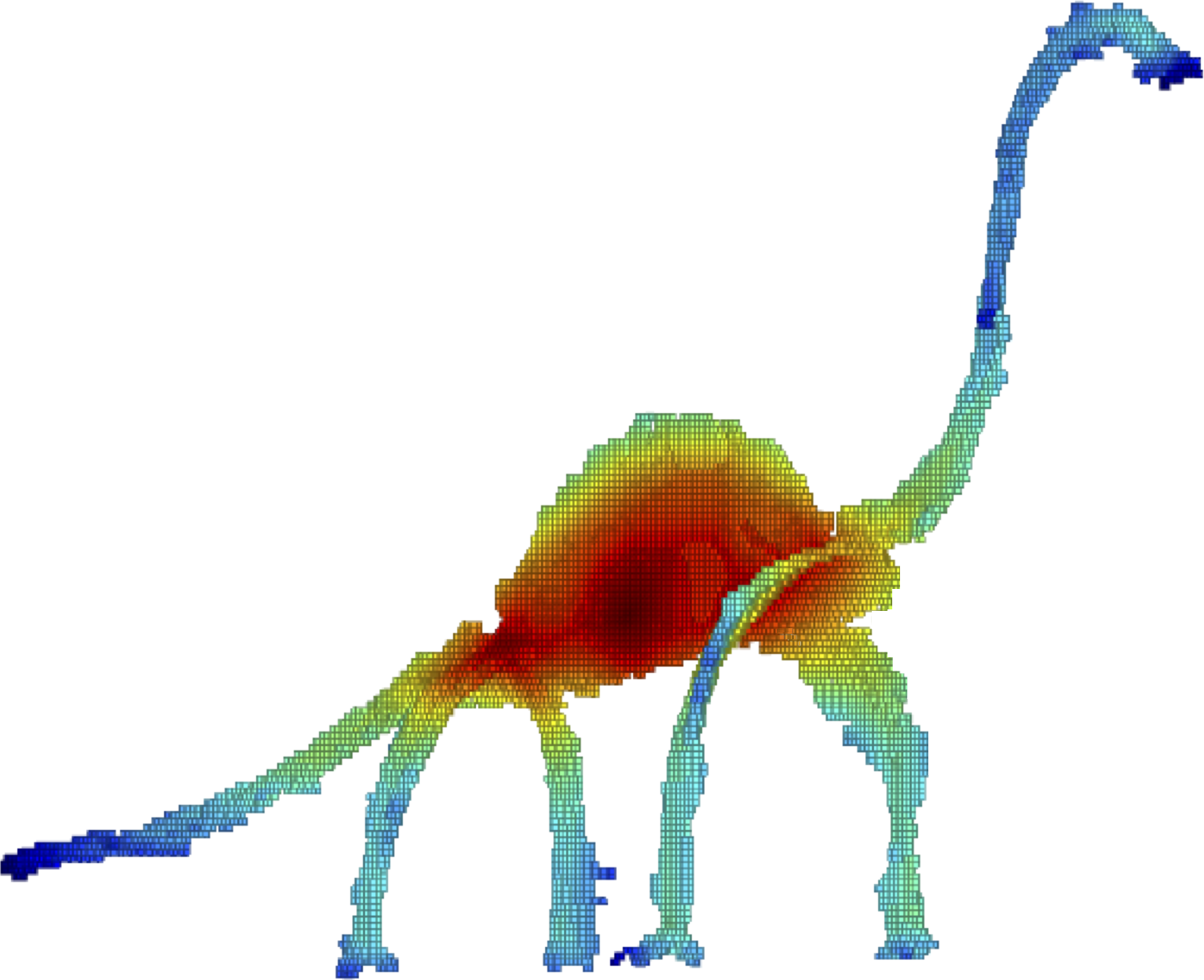}\\
    \textbf{(a)} 3D Shape & \textbf{(b)} $D_E(\mathbf{p})$ & \textbf{(c)} $\frac{\int_{\partial R} \langle \dot{q}, \mathcal{N}_0 \rangle \partial \mathcal{S}}{Area(\partial R)}$ & \textbf{(d)} MAT
\end{tabular}
	\caption{Using the divergence theorem, medial axis voxels can be identified by considering the behavior of the average outward flux (AOF) of the gradient of the Euclidean distance function to the boundary of a 3D object, through a shrinking sphere \cite{siddiqi2008medial}. In particular, the limiting AOF value of all points not located on the skeleton is equal to zero. Starting with the boundary of a 3D object (\textbf{a}), we first compute the distance map to the boundary (\textbf{b}), and then compute the 3D AOF 3D map (\textbf{c}) from this distance. Finally, by retaining the non-zero AOF voxels, we obtain the 3D medial axis transform (\textbf{d}). Since at each medial voxel the maximal inscribed sphere (Fig. \ref{fig:sphere}) touches the boundary at two distinct points, it is possible to reconstruct the boundary purely from the medial locus.
}
	\label{fig:medial_axis_generation}
\end{figure*}

To provide the intuition behind this algorithm consider the object shown in Figure  \ref{fig:sphere}, together with a local portion of its medial surface (light green) and the associated boundary surface patches (purple). The medial surface is homotopic to the original object and thus reflects its topology while making local reflective symmetries explicit. Let $\mathcal{S}$ represent the closed surface of a 3D object and $D_E(\mathbf{p})$ be the Euclidean distance to the boundary of this object at each point $\mathbf{p}$ in its interior.  Now consider the gradient vector field of $D_E(\mathbf{p})$ as $\dot{q}$. Where there is just one closest point $\mathbf{b} = (b_x,b_y,b_z)$ on the boundary, to a particular interior-point $\mathbf{a} = (a_x,a_y,a_z)$, the gradient vector at $\mathbf{a}$ is computed as:
\begin{equation}
    \dot{q}(\mathbf{a}) = \frac{\mathbf{a}-\mathbf{b}}{||\mathbf{a}-\mathbf{b}||}
\end{equation}
However, the gradient vector field is multi-valued at locations on the medial surface.
Exploiting this property, an ``average outward flux'' measure was proposed in \cite{siddiqi2002hamilton}, which computes the outward flux of $\dot{q}$ through a shrinking sphere averaged over the surface area of that sphere:
\begin{equation}
\label{eq:AOF}
    AOF(\mathbf{p}) = \frac{\int_{\partial R} \langle \dot{q}, \mathcal{N}_0 \rangle \partial \mathcal{S}}{Area(\partial R)}.
\end{equation}
Here $\partial R$ represents the surface area of the shrinking sphere and $\partial \mathcal{S}$ is the surface area element. $\mathcal{N}_0$ is the outward normal at each point on the sphere. Earlier work on medial axis computations has shown that as the size of this sphere shrinks to zero, this measure gives non-zero values on the medial surface, but zero-values off the medial locus, and hence provides an effective means for localizing medial surface points \cite{dimitrov2003flux, siddiqi2002hamilton}. 
In the present article, we use a robust implementation of this algorithm. Figure \ref{fig:medial_axis_generation} illustrates the sequence of steps for 3D medial surface generation. 

\subsection{Preserving object topology}
Whereas in a continuous setting one can associate medial surface points with locations where the AOF is non-zero, in a discrete setting care has to be taken to ensure that the resultant set of voxels is homotopic to the original object.
We follow past work \cite{bouix2005hippocampal}, which rests on a definition of endpoints and simple points on a digital 3D lattice:
\begin{enumerate}
    \item A {\em simple point} is a point that cannot be removed without changing the topology of the object. Its removal will either disconnect the object or create a hole or a cavity.
        \item Consider a plane that passes through a point $\mathbf{p}$ such that its intersection with the 3D object results in an open curve. If this curve ends at $\mathbf{p}$, then  $\mathbf{p}$ is an {\em endpoint} of that 3D curve. Additional examples of endpoints include when $\mathbf{p}$ is on the rim or corner of a 3D surface. 
\end{enumerate}
On a 3D digital lattice we can classify both simple points and endpoints by considering the 26 neighborhoods of a particular voxel (the cubic lattice of $3 \times 3 \times 3$). We implement these two notions of simple points and endpoints according to the characterization of \cite{pudney1998distance}, which is based upon different sets of neighbors that share either a point, an edge, or a face with the considered voxel. 

\begin{algorithm}[!b]
\caption{Medial Surface Generation Procedure}\label{alg:cap}
\begin{algorithmic}
\Procedure{Average Outward Flux Computation}{}
\State Compute the Euclidean Distance Function $D_E(\mathbf{p})$
\State Compute the gradient vector field $\nabla D_E$
\State Compute the average outward flux of $\nabla D_E$ using \ref{eq:AOF} 
\EndProcedure
\Procedure{ AOF Topology Preserving Thinning}{}
\State $\text{Heap} \rightarrow H$, Threshold $\rightarrow \tau$
\For{each point $\mathbf{p}$ on the boundary of the object }
\If{$\mathbf{p}$ is simple}
\State insert point $\mathbf{p}$ into $H$ with key value $AOF(\mathbf{p})$
\EndIf
\EndFor 
\While{$H$ is not empty}
\State $\mathbf{p}$ = HeapExtractMax($H$)
\If{$\mathbf{p}$ is simple}
\If{$\mathbf{p}$ is end point and $AOF(\mathbf{p}) > \tau$}
\State Label $\mathbf{p}$ as a medial surface point
\Else
\State Remove $\mathbf{p}$
\EndIf
\EndIf
\For{all neighbors $\mathbf{q}$ of $\mathbf{p}$}
\If{if $\mathbf{q}$ is simple}
inset point $\mathbf{q}$ into $H$
\EndIf
\EndFor
\EndWhile
\EndProcedure
\end{algorithmic}
\end{algorithm}

Algorithm 1 describes the resultant medial surface generation method, which combines the AOF measure with a topology preserving thinning approach, using the above characterization of simple points and endpoints. Figure \ref{fig:medial_axis_generation} provides several examples for a variety of 3D objects. At the end of this process the surface of each 3D object $\mathcal{S}$ can be represented by a set of skeletal points where:
\begin{equation}
 \mathcal{S} = \bigcup_{i = 1}^{n} \mathbf{p}_i^{sk} = \bigcup_{i = 1}^{n} (x_i,y_i,z_i,r_i,\lambda_i).
\end{equation}
Here the triplet $(x_i,y_i,z_i)$ represents the location of the surface point $\mathbf{p}_i$, $r_i$ is the radius of the maximal inscribed sphere touching the object at $\mathbf{p}_i$, and $\lambda_i$ is the average outward flux value at the medial surface point at the center of that sphere. As shown in \cite{siddiqi2008medial} the AOF value also reveals the object angle, and thus provides additional useful object and symmetry information. In the next subsection, we will demonstrate that this representation can be used for object reconstruction.
\begin{figure}[!t]
\centering
\begin{tabular}{ccccc}
\includegraphics[height=0.265\textwidth]{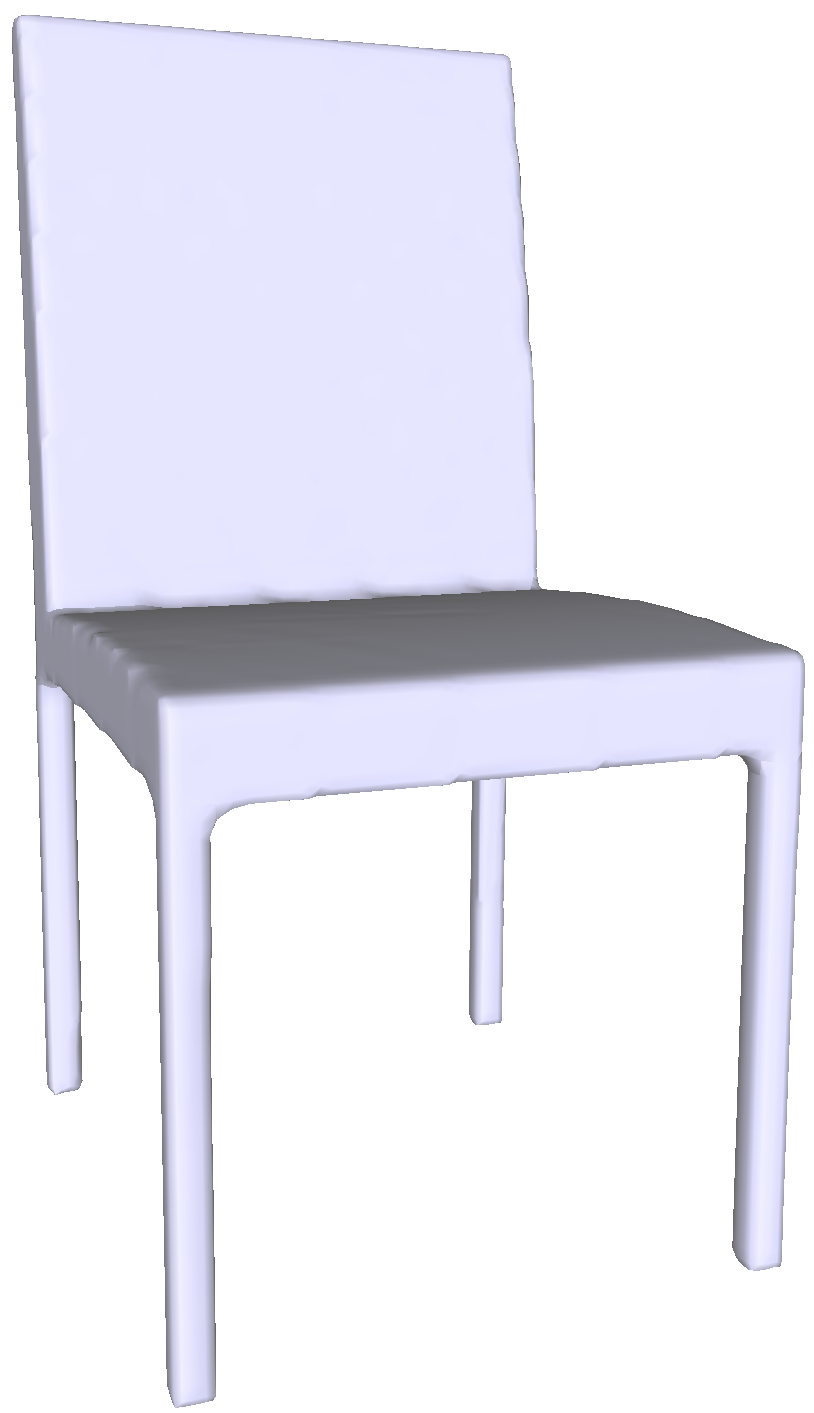} & 
    \includegraphics[height=0.265\textwidth]{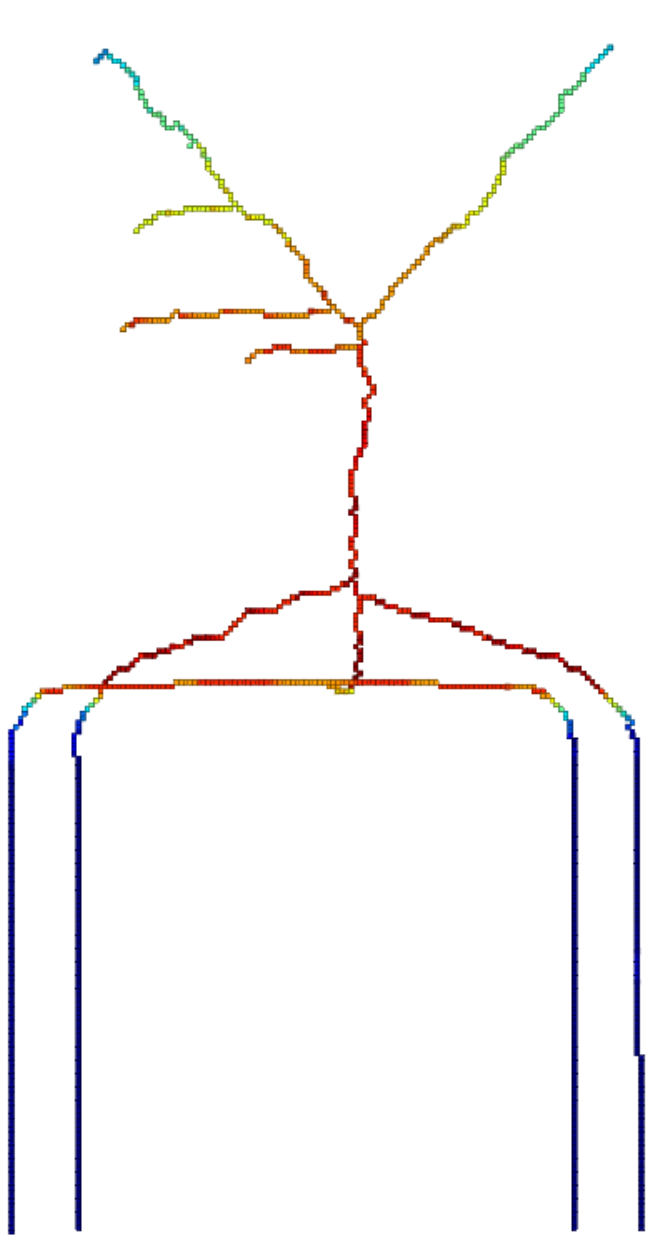} & \includegraphics[height=0.265\textwidth]{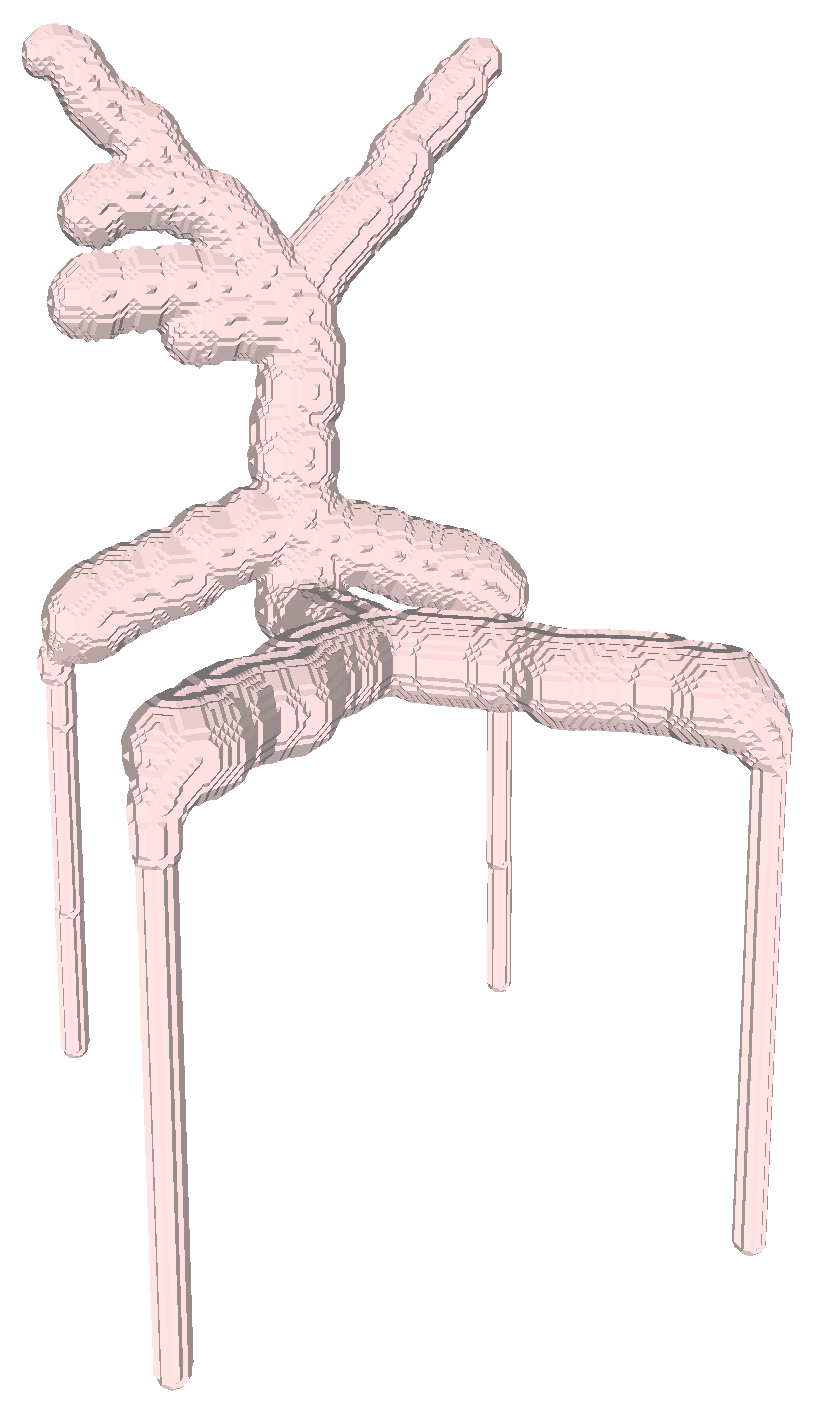} & \includegraphics[height=0.265\textwidth]{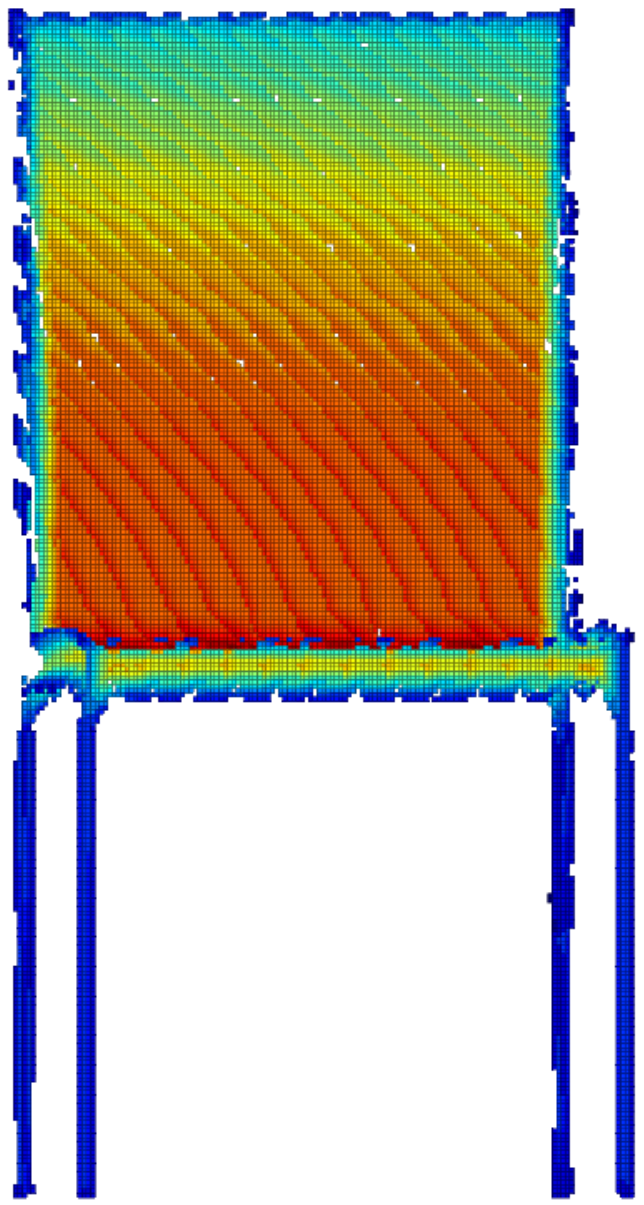}  &  \includegraphics[height=0.265\textwidth]{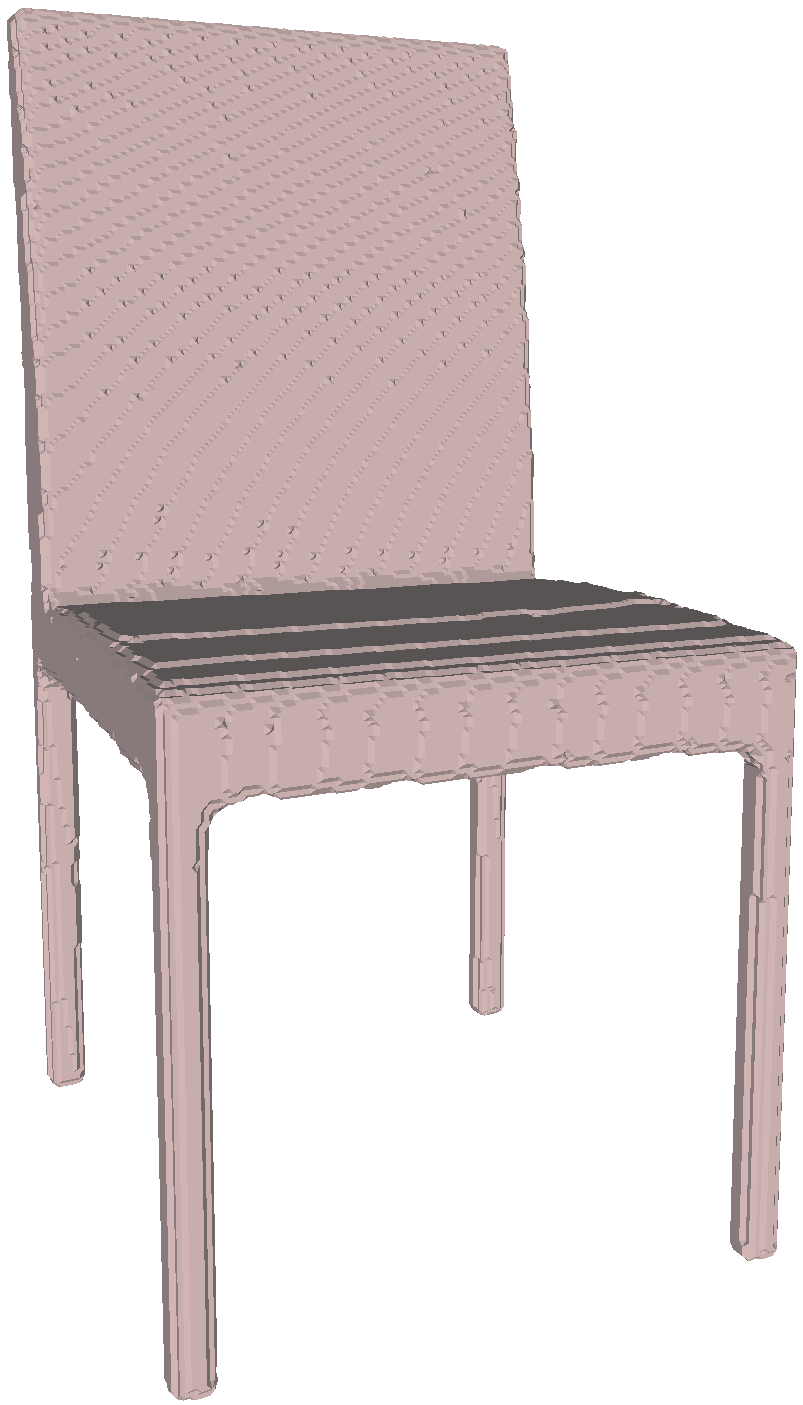}\\
    \multicolumn{1}{c}{\textbf{(a)}} & \multicolumn{2}{c}{\textbf{(b)}} & \multicolumn{2}{c}{\textbf{(c)}}
\end{tabular}
    \caption{\textbf{(a)} The original object. \textbf{(b)} Its skeleton and subsequent reconstruction using the method of \cite{lee1994building}.
    \textbf{(c)} Its AOF based medial surface and subsequent reconstruction.}
	\label{fig:reconstruction_comparison}
\end{figure}

\subsection{Medial Spectral Coordinates for 3D Shape Analysis}
\label{sec:shape_recon}

Perhaps the most obvious criterion for the suitability of an object shape representation method is whether it can be used to reconstruct the original object. This criterion, alongside other criteria such as completeness, hierarchy, invariance, stability, and similarity has been used to judge shape representation choices. 
Whereas medial representations satisfy several of these criteria \cite{siddiqi2008medial}, algorithms implemented on a discrete lattice may not allow for faithful object reconstruction. Here, for purposes of reconstruction, we compare the AOF skeletonization process against a commonly used skeletonization method that extracts 3D skeletons by thinning to produce medial curves \cite{lee1994building}. Since the medial surface consists of the locus of maximally inscribed spheres, the original object can be reconstructed by considering the envelope of the maximally inscribed spheres at all the medial surface points. Figure \ref{fig:reconstruction_comparison} compares reconstructions obtained by these two approaches.
Here we see that for the chair examples, the 3D object reconstructed from  the method in \cite{lee1994building} (Figure \ref{fig:reconstruction_comparison} bottom) is far from satisfactory and important object details are lost, in contrast to the reconstruction obtained from AOF medial surfaces (Figure \ref{fig:reconstruction_comparison} middle). 
A quantitative comparison via a mean intersection over union (mIoU) measure with original objects  is shown in Table \ref{tab:rec_table_scores}, over the entire datasets of ModelNet10 \cite{wu20153d} and COSEG \cite{wang2012active}.

\begin{table}[!b]
    \centering
    \begin{tabular}{|c|c|c||c|c|c|}
        \hline
        \multicolumn{3}{|c||}{COSEG \cite{wang2012active}} & \multicolumn{3}{|c|}{ModelNet10 \cite{wu20153d}}\\\hline\hline
        {\scriptsize Model } & {\scriptsize Ours} & {\scriptsize Curve} & {\scriptsize Model } & {\scriptsize Ours} & {\scriptsize Curve} \\\hline
        {\scriptsize Candelabra} & 96.99 & 72.82 & {\scriptsize Table} & 99.68 & 88.61\\\hline
        {\scriptsize Chairs} & 96.38 & 81.40 & {\scriptsize Bathtub} & 93.25 & 86.50\\\hline
        {\scriptsize Fourleg}	& 90.33 & 72.44 & {\scriptsize Bed} & 95.31 & 73.13\\\hline
        {\scriptsize Goblets} & 90.68 & 81.69 & {\scriptsize Chair} & 96.10 & 85.04 \\\hline
        {\scriptsize Guitars}	& 93.19 & 75.42 & {\scriptsize Desk} & 97.78 & 80.73\\\hline
        {\scriptsize Lamps}	& 95.30 & 66.71 & {\scriptsize Dresser} & 94.23 & 91.31 \\\hline
        {\scriptsize Vases}	& 96.54 &69.51  & {\scriptsize Monitor} & 90.90 & 73.02\\\hline
        {\scriptsize Irons} & 94.07 & 80.43&  {\scriptsize Night stand} & 92.66 & 82.24 \\\hline
        {\scriptsize Tele-aliens}	& 98.19 &  75.46& {\scriptsize Sofa} & 91.53 & 81.48\\\hline
        {\scriptsize Chairs (L)} &  97.18 & 79.05 & {\scriptsize Toilet} & 91.05 & 82.21\\\hline\hline
        {\scriptsize Mean} &  94.89 & 75.49 & {\scriptsize Mean} & 94.25 & 82.43\\
         \hline
    \end{tabular}
    \caption{The mean Intersection-Over-Union (mIoU) reconstruction measures, for the two datasets of COSEG and ModelNet10, comparing the use of medial curves \cite{lee1994building} and AOF medial surfaces \cite{siddiqi2002hamilton}.}
    \label{tab:rec_table_scores}
\end{table}

\section{Spectral Coordinates using Medial Manifolds}
\label{sec:spectral_coordinate}
We now propose to use the medial manifold to equip the object's surface with a new spectral signature, one that explicitly considers local object width. Whereas spectral approaches have been popular for shape correspondence \cite{bronstein2010shape,lombaert2012focusr,bronstein2017geometric,jain2006robust} and spectral methods have also been successfully used in combination with graph convolution to learn features from point sets and meshes \cite{wang2018local, yi2017syncspeccnn, donati2020deep}, the explicit consideration of object width and the associated symmetry properties and coupling of surface points which share a medial ball is new. 

\begin{figure*}[!t]
    \centering
    \includegraphics[width = 1.04\textwidth]{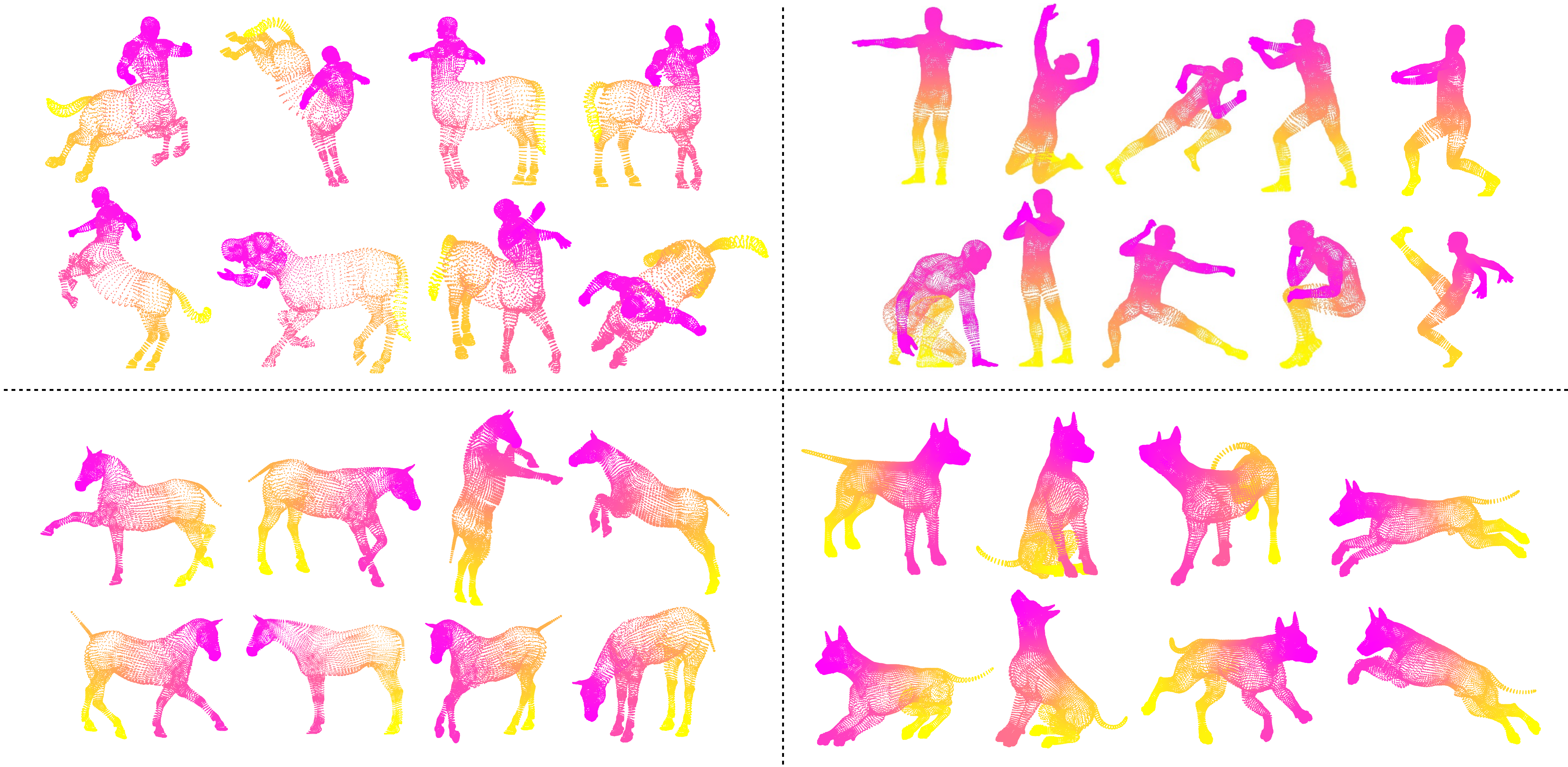}
    \caption{Qualitative results showing correspondences obtained using our medial spectral coordinates across different poses of the same models in the TOSCA \cite{bronstein2008numerical} dataset.}
    \label{fig:corr_qualitative}
\end{figure*}

Let the object's surface $\mathcal{S}$ be represented by its set of boundary points $\{\mathbf{b}_1,\mathbf{b}_2,\dots,\mathbf{b}_n\}$. We voxelize the object and compute the medial surface using Algorithm 1 in Section \ref{sec:medialaxis}. Let the set of resulting medial surface voxels be $\{\mathbf{p}_1^{sk}, \mathbf{p}_2^{sk}, \dots, \mathbf{p}_m^{sk}\}$. Using the reconstruction algorithm discussed in Section \ref{sec:shape_recon}, we reconstruct a boundary surface $\mathcal{S}^{r}$ that is close to $\mathcal{S}$, the surface of the original object. Using a $kd$-tree partitioning algorithm, we then find a correspondence map between points from $\mathcal{S}$ to $\mathcal{S}^{r}$. Using this map, we associate each point $\mathbf{b}_i$ on the boundary with its corresponding medial locus point $\mathbf{p}_j^{sk}$ on the medial surface. Now, assuming that the boundary is tessellated so that the boundary points are the vertices (or set of nodes) in a connected graph, eigenmaps of these nodes are computed as follows. We first compute an adjacency matrix (as shown in Figure \ref{fig:medial_axis_generation} (\textbf{d})) where a particular metric is used to compute the adjacency weights in the matrix $W$, as follows.
Consider two boundary points (nodes) $\mathbf{b}_i$ and $\mathbf{b}_j$. Let these boundary points be associated with skeletal points $\mathbf{p}_i$ and $\mathbf{p}_j$, respectively. Now, let the maximal inscribed sphere at skeletal point $\mathbf{p}_i$ be given by $\rho(\mathbf{p}_i^{sk})$. We consider the degree of overlap between the inscribed spheres touching the object's surface at $\mathbf{b}_i, \mathbf{b}_j$ to construct the adjacency weight between the associated nodes
\begin{equation}
\label{eq:adj_weight}
    w_{ij} = \frac{C(i,j)}{\rho(\mathbf{p}_i^{sk})},
\end{equation}
where $C(i,j)$ is the volume of the intersection between the spheres $\rho(\mathbf{p}_i^{sk})$ and $\rho(\mathbf{p}_j^{sk})$. We then
construct the general Laplacian operator as $L = D-W$, where $W$ is the weighted adjacency matrix of the graph with affinity weights (see \cite{grady2010discrete}), and the degree matrix, $D$, is a diagonal matrix, where $D_{ii} = \Sigma_j w_{ij}$. We then compute the eigenvectors of the Laplacian matrix:
\begin{equation}
\label{eq:eigen_vec}
    LE_{\lambda_i} = \lambda_i E_{\lambda_i}, i =  1,\dots,n.
\end{equation}
Here, since matrix $W$ is not symmetric (leading to a possibility of imaginary numbers numerically), we create two symmetric matrices $W^{sym}$ and $D^{sym}$ as follows. $W^{sym}$ has entries given by
\begin{equation}
    w_{ij}^{sym} = C(i,j),
\end{equation}
and $D^{sym}$ which is a diagonal matrix, with entries $d_{ii}^{sym} =\rho(\mathbf{p}_i^{sk})$. Now, by using the Arnoldi method of ARPACK \cite{lehoucq1998arpack}, we can obtain eigenvectors of the following equation:
\begin{equation}
    W^{sym}E_{\lambda_i} = \lambda_i D^{sym} E_{\lambda_i}, i = 1,\dots,n
\end{equation}
where the  obtain eigenvalues are real and non-negative. Now that we have our eigenvectors, we can obtain spectral coordinates by mapping the coordinates from the boundary of the 3D shape. For each boundary point $\mathbb{b}_i$ element (e.g. $x$, $y$, $z$ or its corresponding sphere radius value $r$ from the medial surface), we can compute the mapped information on the $i^{th}$ basis eigenvectors as:
\begin{equation}
\mathbf{m}_i = \mathbf{b}_i D^{sym} E_{\lambda_i}.
\end{equation}
Referring back to the Dirichlet energy equation introduced by Bronstein \textit{et al.} \cite{bronstein2017geometric} that measures change over the boundary $\mathcal{S}$, we can introduce our spectral coordinates as follows:
\begin{equation}
S_C(\mathcal{S}) = \Big(\bigcup_{i = 1}^{n} \lambda_i \sum_{j = 1}^k(\mathbf{m}_{ij}^2)\Big).
\end{equation}
We are careful to always remove the trivial eigenvalue of 0 (the smallest eigenvalue) from our set of spectral coordinates. 

\section{Applications}
\label{sec:applications}
We now evaluate the use of our medial  spectral coordinates for three popular applications in 3D shape analysis:
a) finding correspondence between surface points, b) 3D object part segmentation of 3D objects and c) 3D object classification. For each case, we demonstrate that the explicit use of local object width, as reflected in the medial radius, and the coupling of surface points that share a common medial sphere, provides direct benefits and improved performance.

\subsection{Shape Correspondence}
\begin{figure}
    \centering
    \includegraphics[width = 0.45\textwidth]{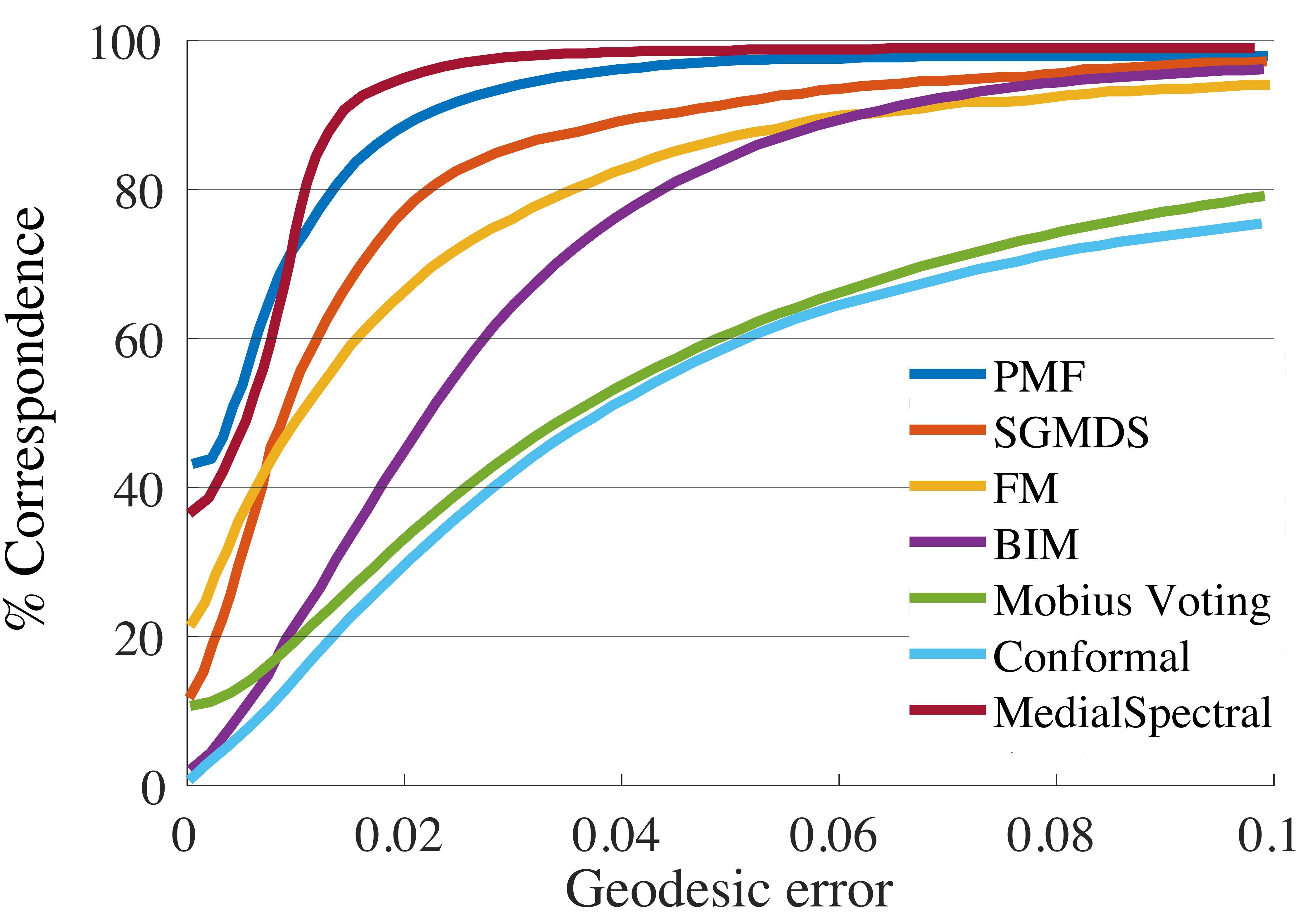}
    \caption{Correspondence accuracy on the TOSCA dataset \cite{bronstein2008numerical} for different methods. The methods we have compared include PMF \cite{lahner2017efficient}, SGMDS \cite{aflalo2016spectral}, FM \cite{ovsjanikov2012functional}, BIM \cite{kim2011blended}, Mobius Voting \cite{lipman2009mobius}, Conformal \cite{kim2011blended}, and the use of our new MedialSpectral signature.}
    \label{fig:tosca_result}
\end{figure}

To obtain correspondences between surface points on two distinct objects, one needs a matching algorithm that matches one against the other. In the present setting, we have spectral coordinates alongside eigenvalues that are obtained from a Laplacian operator. 
When spectra are computed, two situations are possible that make the direct comparison of spectral coordinates challenging. First, eigenvector computation may generate a sign ambiguity. Second, it is possible that when eigenvectors are being computed for the same data but for two different shapes, they might be computed in opposite orders due to the fact that the ordering of the lowest eigenvector may change.  \cite{lombaert2012focusr} suggests a method for mitigating the effects of this flipping problem by favoring three factors: 1) pairs of eigenvectors that are most likely to match based on the similarity between their eigenvalues 2) histograms 3) the spatial distributions of their spectral coordinate value. We use these techniques to reorder and align spectra during the correspondence finding process. The process of reordering is sped up by downsampling all eigenvectors. After reordering and aligning the spectra, two points that are closest in the embedded representations can be treated as corresponding points across the two objects. This is achieved by using the Coherent Point Drift (CPD) method \cite{myronenko2010point}. We test our correspondence finding algorithm on the TOSCA dataset \cite{bronstein2008numerical}, which includes a total of 80 objects, including four limb animals such as cats and dogs, female and male figures. We compare our methods against several other existing methods and report the results in Figure \ref{fig:tosca_result}. The results show the advantages of using medially driven spectral coordinates when compared to the best conformal model that we could find. We also achieve results that are comparable to those produced by the state of the art deep neural network based models, such as PMF \cite{lahner2017efficient}. We show qualitative results of correspondences obtained on four different models from the TOSCA dataset in Figure \ref{fig:corr_qualitative}.

\begin{figure*}[!t]
    \centering
    \begin{tabular}{c@{\hskip 10pt}c@{\hskip 10pt}c}
        \includegraphics[height = 0.26\textwidth]{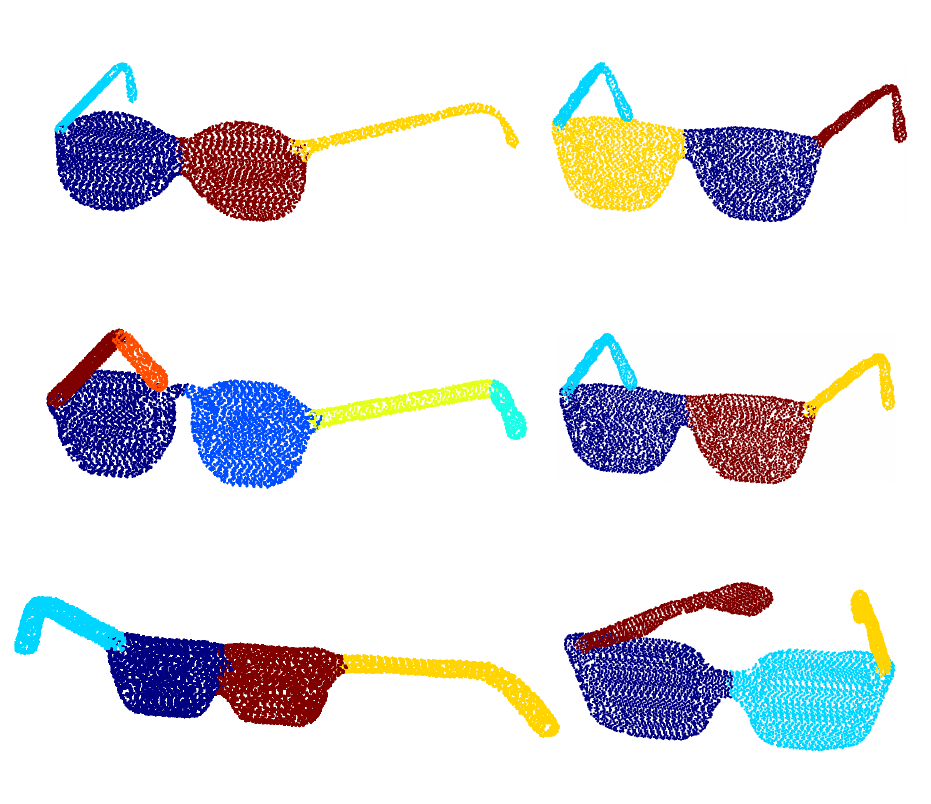} & \includegraphics[height = 0.265\textwidth]{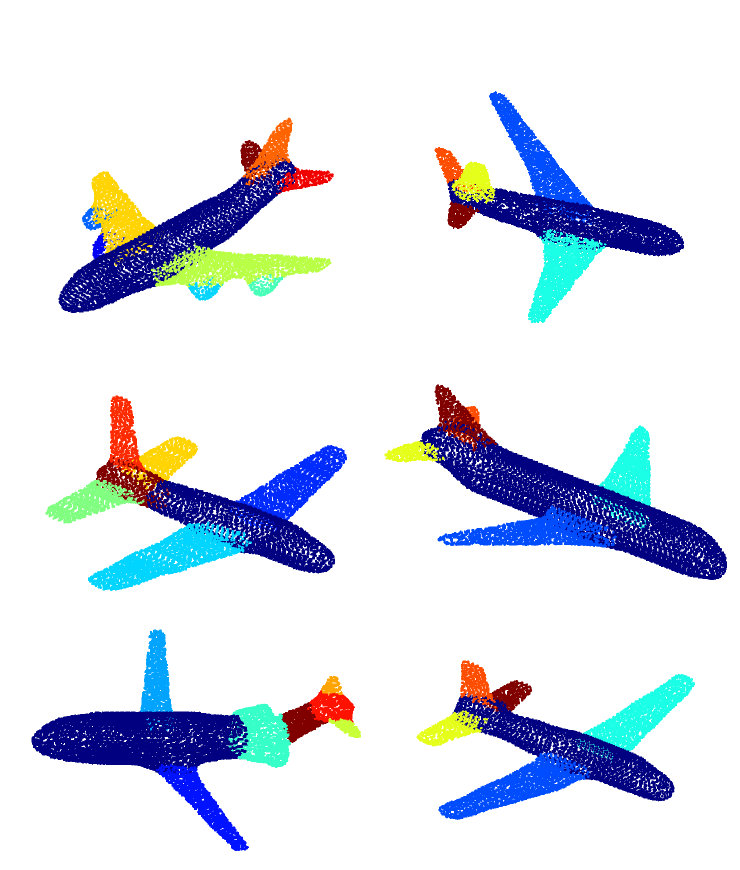} & \includegraphics[height = 0.26\textwidth]{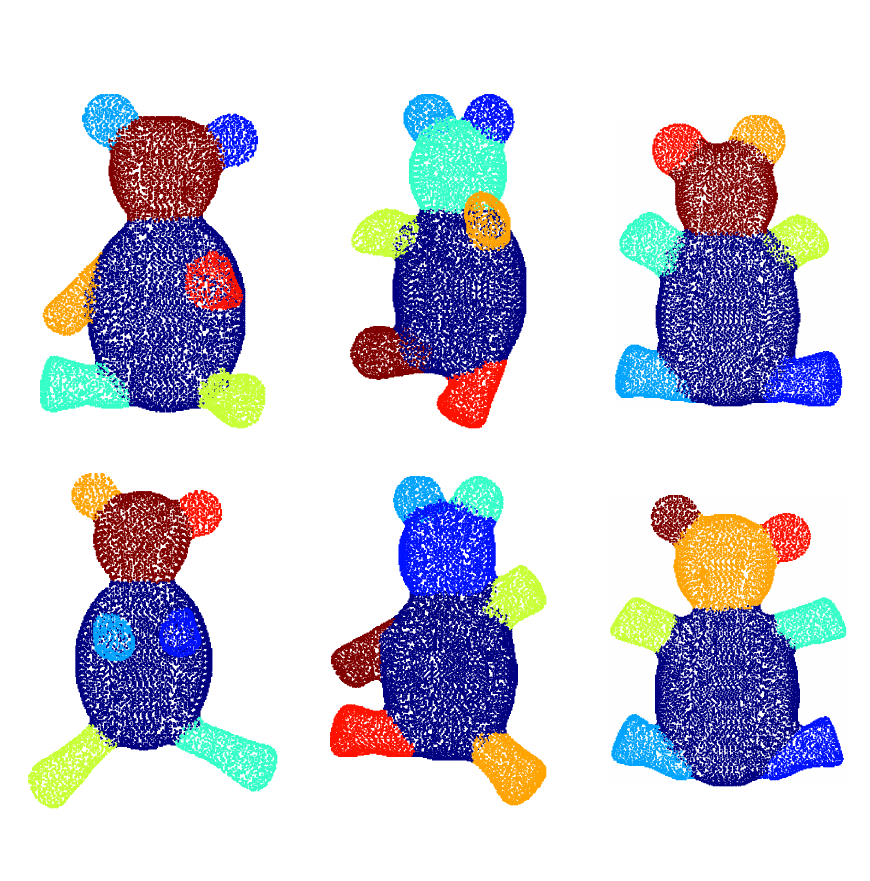}\\ 
        Glasses & Airplane & Teddy\\
        \includegraphics[height = 0.26\textwidth]{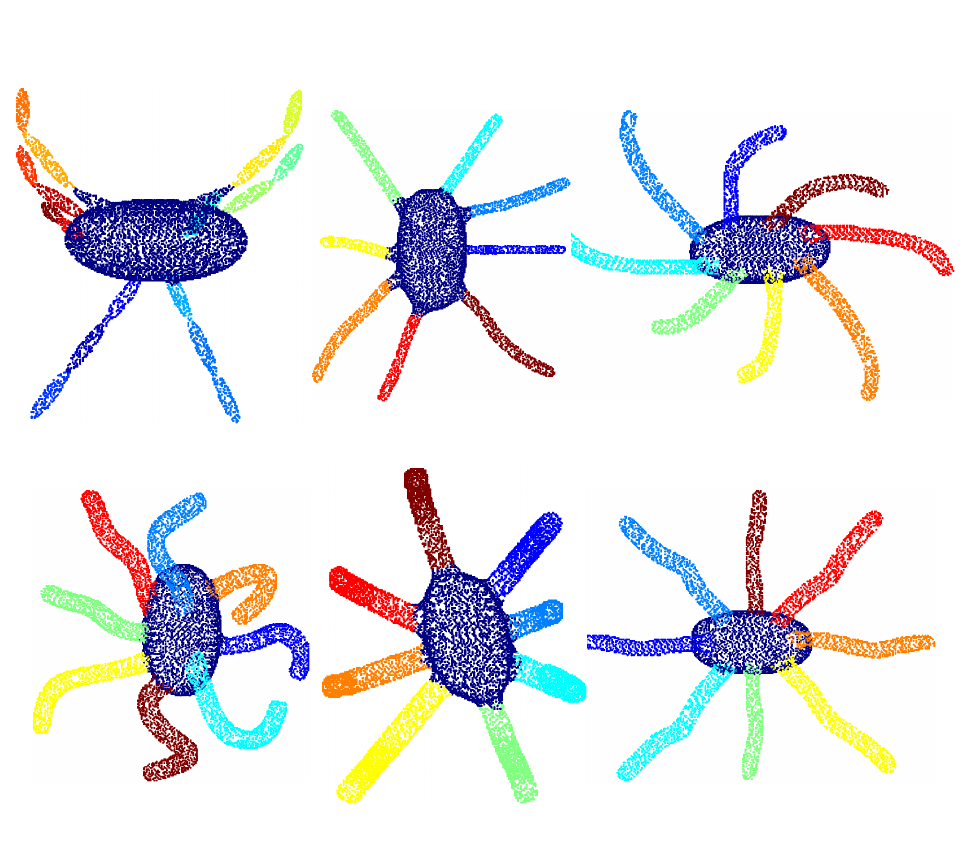} &  \includegraphics[height = 0.26\textwidth]{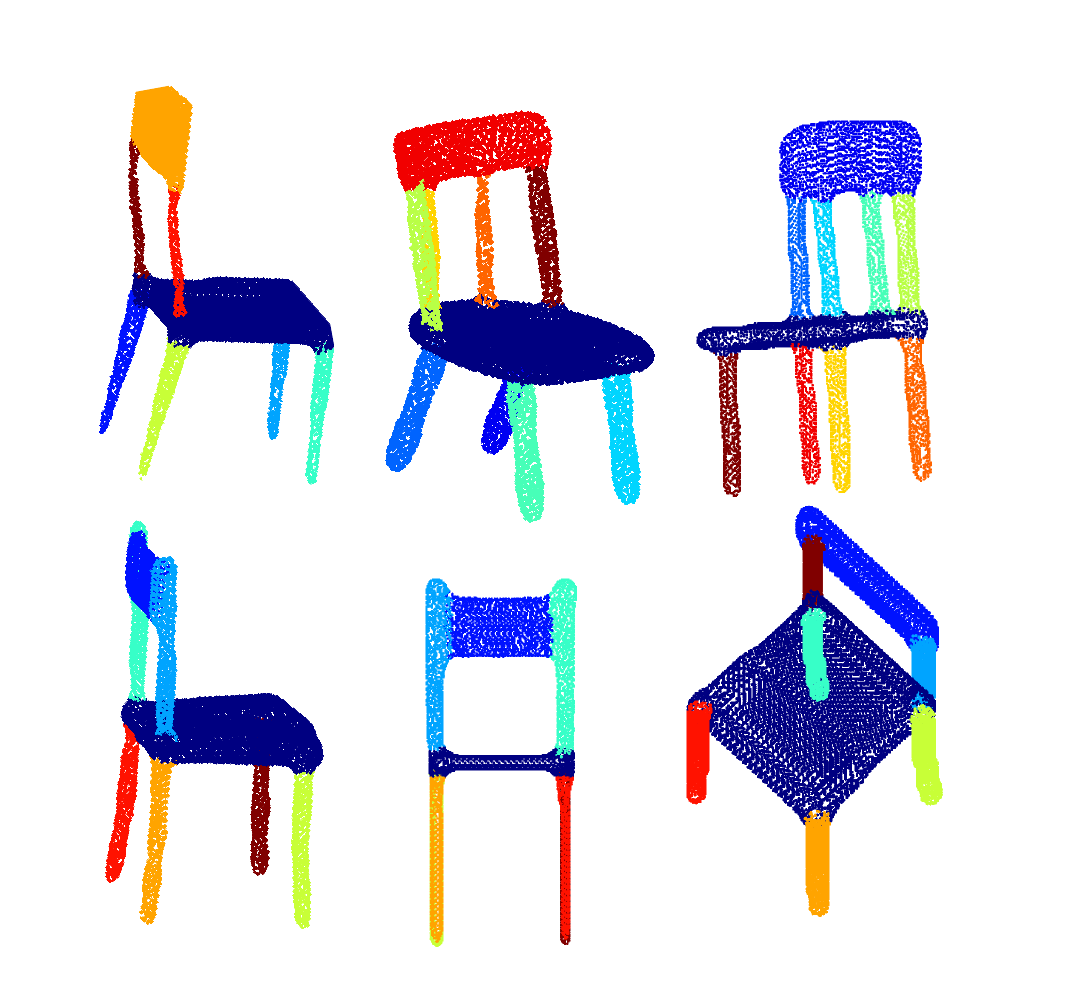} & 
        \includegraphics[height = 0.25\textwidth]{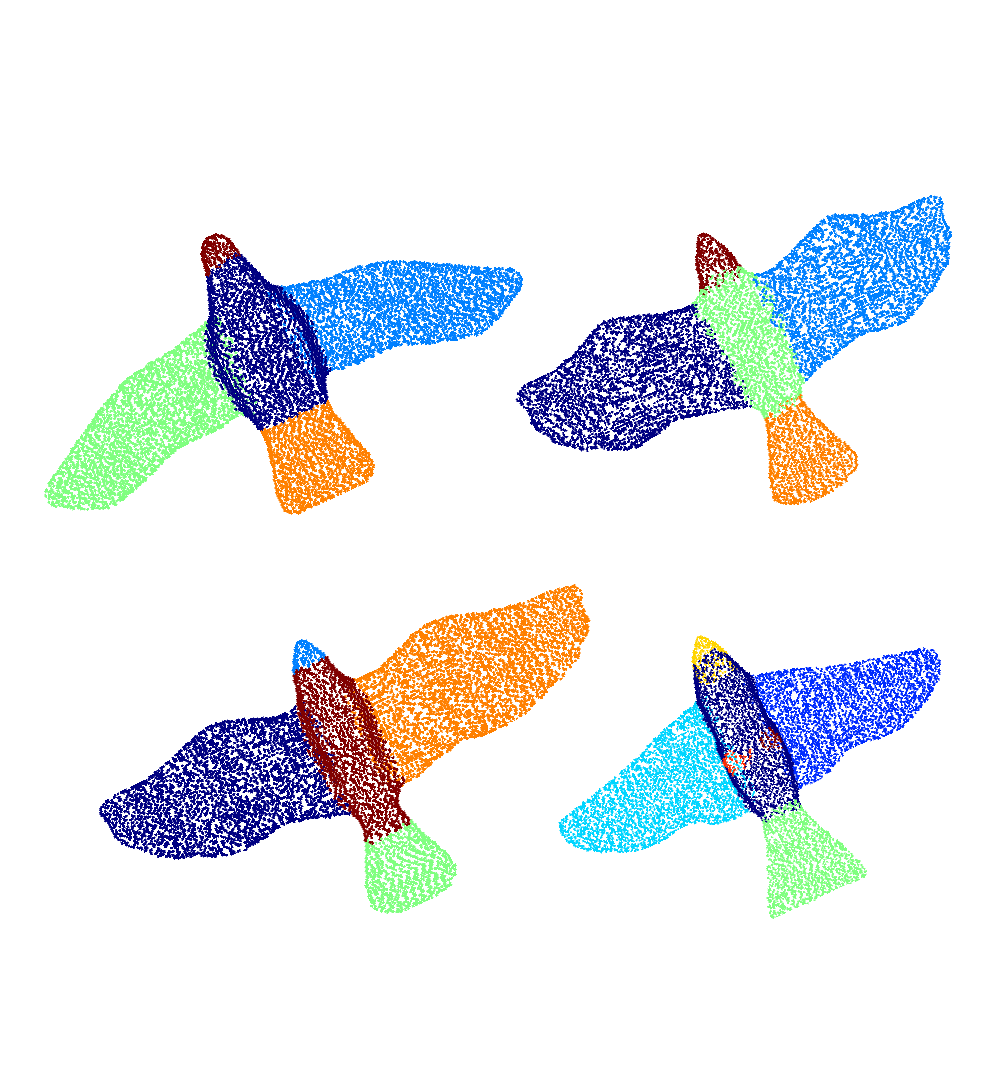} \\
        Octopus & Chair & Bird\\
    \end{tabular}
    \caption{Qualitative part segmentation results obtained using medially driven spectral signatures on the Princeton segmentation benchmark (see text for details).}
    \label{fig:representative_figures_part_segmentation}
\end{figure*}

\begin{table*}[!t]
    \centering
    \begin{tabular}{c@{\hskip 3pt}c@{\hskip 3pt}c@{\hskip 3pt}c@{\hskip 3pt}c@{\hskip 3pt}c@{\hskip 3pt}c@{\hskip 3pt}c@{\hskip 3pt}c@{\hskip 3pt}c@{\hskip 3pt}c@{\hskip 3pt}c@{\hskip 3pt}c}
    \Xhline{2\arrayrulewidth}
    {\scriptsize Model} & {\scriptsize SpectralMedial} & {\scriptsize Spectral} & {\scriptsize MV-RNN \cite{LE2017103}} & {\scriptsize Shu \cite{SHU201639}} & {\scriptsize WcSeg} & {\scriptsize RandCuts} &  {\scriptsize ShapeDiam} & {\scriptsize NormCuts} & {\scriptsize CoreExtra} & {\scriptsize RandWalks} & {\scriptsize FitPrim} & {\scriptsize KMeans}\\\Xhline{2\arrayrulewidth}
{\scriptsize Human}  & \textbf{0.0557} & 0.1240 & 0.106  & 0.116 & 0.128 & 0.131 & 0.179 & 0.152 & 0.225 & 0.219 & 0.153 & 0.163\\\hline
{\scriptsize Cup} & 0.1413 & 0.1262 & 0.100  & \textbf{0.096} & 0.171 & 0.219 & 0.358 & 0.244 & 0.307 & 0.358 & 0.413 & 0.459\\\hline
{\scriptsize Glasses} & 0.0802 & 0.1304 & \textbf{0.066}  & 0.173 & 0.173 & 0.101 & 0.204 & 0.141 & 0.301 & 0.311 & 0.235 & 0.188\\\hline
{\scriptsize Airplane} & \textbf{0.0728} & 0.0891 & 0.085  & 0.150 & 0.089 & 0.122 & 0.092 & 0.186 & 0.256 & 0.248 & 0.166 & 0.211\\\hline
{\scriptsize Ant} & 0.0279 & 0.0900 & 0.021  & \textbf{0.001} & 0.021 & 0.025 & 0.022 & 0.047 & 0.065 & 0.068 & 0.086 & 0.131\\\hline
{\scriptsize Chair} & 0.0483 & 0.0518 & 0.051  & \textbf{0.040} & 0.103 & 0.184 & 0.111 & 0.088 & 0.187 & 0.156 & 0.212 & 0.213\\\hline
{\scriptsize Octopus} & 0.0255 & 0.0835 & \textbf{0.022}  & 0.036 & 0.029 & 0.063 & 0.045 & 0.061 & 0.051 & 0.067 & 0.101 & 0.101\\\hline
{\scriptsize Table} & 0.0630 & 0.1158 & 0.072  & \textbf{0.040} & 0.091 & 0.383 & 0.184 & 0.093 & 0.244 & 0.131 & 0.181 & 0.369\\\hline
{\scriptsize Teddy} & 0.0521 & 0.1105 & 0.035  & \textbf{0.024} & 0.056 & 0.045 & 0.057 & 0.121 & 0.114 & 0.128 & 0.132 & 0.182\\\hline
{\scriptsize Hand} & \textbf{0.0670} & 0.1443 & 0.076  & 0.135 & 0.116 & 0.090 & 0.202 & 0.155 & 0.155 & 0.189 & 0.202 & 0.154\\\hline
{\scriptsize Plier} &  0.0970 & 0.1433 & \textbf{0.054}  & 0.151 & 0.087 & 0.109 & 0.375 & 0.183 & 0.093 & 0.230 & 0.169 & 0.263\\\hline
{\scriptsize Fish} &\textbf{0.1030} & 0.2339 & 0.146  & 0.288 & 0.203 & 0.297 & 0.248 & 0.394 & 0.273 & 0.388 & 0.424 & 0.413\\\hline
{\scriptsize Bird} & \textbf{0.0049} & 0.0850 &  0.059  & 0.171 & 0.101 & 0.107 & 0.115 & 0.184 & 0.124 & 0.250 & 0.196 & 0.190\\\hline
{\scriptsize Armadillo} & \textbf{0.0178} & 0.1265 & 0.060 & 0.073 & 0.081 & 0.092 & 0.090 & 0.116 & 0.141 & 0.115 & 0.091 & 0.117\\\hline
{\scriptsize Bust} & \textbf{0.1081} & 0.1729 & 0.162 & 0.275 & 0.266 & 0.232 & 0.298 & 0.316 & 0.315 & 0.298 & 0.300 & 0.334\\\hline
{\scriptsize Mech} & 0.1018 & 0.1133 & 0.121  & \textbf{0.073} & 0.182 & 0.277 & 0.238 & 0.159 & 0.387 & 0.211 & 0.306 & 0.425\\\hline
{\scriptsize Bearing} &0.0660 & 0.1516 & 0.080 & \textbf{0.056} & 0.122 & 0.124 & 0.119 & 0.183 & 0.398 & 0.246 & 0.188 & 0.280\\\hline
{\scriptsize Vase} & \textbf{0.0772} &  0.1504 & 0.106  & 0.212 & 0.161 & 0.133 & 0.239 & 0.236 & 0.226 & 0.246 & 0.257 & 0.387\\\hline
{\scriptsize FourLeg} &\textbf{0.0921} &  0.1772 & 0.135  & 0.140 & 0.152 & 0.174 & 0.161 & 0.208 & 0.191 & 0.218 & 0.185 & 0.193\\\Xhline{2\arrayrulewidth}
{\scriptsize Average} & \textbf{0.0696} & 0.1273 &  0.082  & 0.118 & 0.123 & 0.153 & 0.176 & 0.172 & 0.211 & 0.215 & 0.210 & 0.251\\\Xhline{2\arrayrulewidth}

\end{tabular}
    \caption{The Rand Index (error) segmentation scores, for object categories in the Princeton segmentation benchmark, comparing across different methods. In this table, a lower number indicates better performance. The use of medially drive spectral coordinates (column two) consistently outperforms the use of spectral coordinates without the medial (radius) component (column three). The approach is also competitive against many other approaches for several of the object categories.}
    \label{tab:part_segmentation_result}
\end{table*}
\subsection{Part Segmentation}
We now examine whether the addition of medially driven spectral coordinates can be beneficial to the task of part segmentation. To achieve this aim, we propose the following task to carry out unsupervised 3D shape decomposition. We collect all the spectral coordinates computed and add them as additional features to the boundary representation $\mathbf{b}_i = (x_i,y_i,z_i,r_i)$. We then use the high-dimensional data spectral clustering of \cite{cai2020spectral} to cluster all object models within a category together. The algorithm of \cite{cai2020spectral} provides a spectral clustering approach based on subspace randomization and graph fusion for high-dimensional data, which enables us to carry out both segmentation and co-segmentation tasks for a single shape or family of shape models (co-segmentation). To examine whether these added features to the original shape help with the segmentation task, we tested our off-the-shelf method on the Princeton Segmentation Benchmark \cite{Chen:2009:ABF}. We report our results in Table \ref{tab:part_segmentation_result} and show typical qualitative part segmentation results in Figure \ref{fig:representative_figures_part_segmentation}. 
 
\subsection{Object Classification}
The last set of experiments we present in this paper relate to the problem of object classification. For this task, we examine whether the inclusion of medial spectral coordinates leads to an improvement in the performance of a vanilla neural network model. To achieve this goal, we carried out the following classification task. Inspired by the idea presented in \cite{xu2020geometry},  we created a new feature vector for each point in a 3d point cloud as a Geometry Similarity Connection, capturing the local behavior of each point in the eigenvector space. First, we find the $k$ nearest neighbors to each point in the eigenvector space. Then, we add the mean and standard deviation of those $k$ points to the $x$,$y$,$z$ coordinates of the considered point. Finally, we add these additional features to the point's features, creating a vector of $9$ features for each point in the point cloud. We tested the added features on the popular PointNet model \cite{qi2017pointnet}. The added features result in an improvement in the overall accuracy of the model from 88.76\% to 90.41\% on ModelNet-10 and from 86.23\% to 88.26\% on ModelNet-40. We present a detailed account of the results for each category for this experiment in Table \ref{tab:PoinNET_AND_EIGENS}. 

\begin{table*}[!b]
    \centering
    \begin{tabular}{c@{\hskip 20pt}c@{\hskip 10pt}c@{\hskip 10pt}c@{\hskip 10pt}c@{\hskip 10pt}c@{\hskip 10pt}c@{\hskip 10pt}c@{\hskip 10pt}c@{\hskip 10pt}c@{\hskip 10pt}c@{\hskip 10pt}c}
      \\\Xhline{2\arrayrulewidth}

\multirow{ 3}{*}{\rotatebox[origin=r]{90}{\scriptsize MNet10 \hspace{-3mm}}} & & {\scriptsize	 bathtub }&{\scriptsize	 bed }&{\scriptsize	 chair }&{\scriptsize	 desk }&{\scriptsize	 dresser }&{\scriptsize	 monitor }&{\scriptsize	 night stand }& {\scriptsize	  sofa }&{\scriptsize	 table }&{\scriptsize	 toilet }\\\cline{2-12}
& {\scriptsize	Vanila} & 83.87 & 92.85 & 94.73 & 77.77 & 83.87 & 96.48 & 84.02 & 91.70 & 80.78 & 95.87\\
& {\scriptsize	Ours} & 85.71 & 94.28 & 92.85 & 82.80 & 85.54 & 96.51 & 84.33 & 96.44 & 83.40 & 97.02\\\Xhline{2\arrayrulewidth}

{\scriptsize \multirow{ 9}{*}{\rotatebox[origin=r]{90}{MNet40\hspace{1cm}}}} & &  {\scriptsize	 airplane }&{\scriptsize	 bathtub }&{\scriptsize	 bed }&{\scriptsize	 bench }&{\scriptsize	 bookshelf }&{\scriptsize	 bottle }&{\scriptsize	 bowl }&{\scriptsize	 car }&{\scriptsize	 chair }&{\scriptsize	 cone}\\\cline{2-12}
& {\scriptsize	Vanila} & 100.00 &79.86 &93.88 &74.93 &92.81 &93.98 &100.00 &97.77 &95.83 &100.00\\
& {\scriptsize	Ours} & 100.00 &83.50 &98.19 &76.38 &96.57 &94.67 &100.00 &98.58 &98.88 &100.00
\\\cline{2-12}

& & {\scriptsize	 cup }&{\scriptsize	 curtain }&{\scriptsize	 desk }&{\scriptsize	 door }&{\scriptsize	 dresser }&{\scriptsize	 flower pot }&{\scriptsize	 glass box }&{\scriptsize	 guitar }&{\scriptsize	 keyboard }&{\scriptsize	 lamp}\\\cline{2-12}
& {\scriptsize	Vanila} & 69.85 &89.83 &78.94 &94.89 &64.90 &29.89 &93.93 &100.00 &100.00 &89.85\\
& {\scriptsize	Ours} & 73.95 &92.46 &81.39 &99.29 &66.83 &32.20 &98.71 &100.00 &100.00 &90.93
\\\cline{2-12}

& & {\scriptsize	  laptop }&{\scriptsize	 mantel }&{\scriptsize	 monitor }&{\scriptsize	 night stand }&{\scriptsize	 person }&{\scriptsize	 piano }&{\scriptsize	 plant }&{\scriptsize	 radio }&{\scriptsize	 range hood }&{\scriptsize	 sink}\\\cline{2-12}
& {\scriptsize	Vanila} & 100.00 &95.90 &94.86 &82.41 &84.93 &88.63 &72.85 &69.81 &90.99 &79.93\\
& {\scriptsize	Ours} & 100.00 &98.87 &98.30 &84.37 &88.03 &92.77 &73.09 &70.41 &95.77 &83.19
\\\cline{2-12}

& & {\scriptsize	  sofa }&{\scriptsize	 stairs }&{\scriptsize	 stool }&{\scriptsize	 table }&{\scriptsize	 tent }&{\scriptsize	 toilet }&{\scriptsize	 tv stand }&{\scriptsize	 vase }&{\scriptsize	 wardrobe }&{\scriptsize	 xbox}\\\cline{2-12}
& {\scriptsize	Vanila} & 
95.87 &84.94 &89.95 &87.86 &94.88 &98.88 &86.87 &78.79 &59.93 &69.91\\
& {\scriptsize	Ours} & 97.13 &86.97 &90.60 &89.31 &96.26 &100.00 &87.74 &80.50 &60.59 &74.32
\\\Xhline{2\arrayrulewidth}

    \end{tabular}
    \caption{The difference in object classification accuracy when using vanilla features and our medial spectral features with the PointNet model\cite{qi2017pointnet}, using the ModelNet10 and ModelNet40 datasets \cite{wu20153d}.}
     \label{tab:PoinNET_AND_EIGENS}
\end{table*}
Our results show that in several of these object categories, there is a boost in performance when we use spectral coordinates that use medial surface information over the use of raw data (with the spectral clustering algorithm). These results are reflected in the second (SpectralMedial) and third (Spectral) columns of Table \ref{tab:part_segmentation_result}. We also observe that the use of spectral coordinates outperforms several of the other existing approaches including the deep model-based (\cite{LE2017103}). Finally, the mean performance and the rand index measure of our approach is the best overall on the entire Princeton dataset \cite{Chen:2009:ABF}.
\section{Conclusion}
\label{sec:conclusion}
We have proposed a novel spectral coordinate for 3D shape analysis applications, one that includes the local object width associated with the medial surface radius function. To our knowledge, this is the first attempt to include medial width in a spectral feature, and our experiments demonstrate the considerable benefits of doing this. Our model ties surface points on a 3D object to their associated medial surface points and uses the duality 
between the medial surface of an object and its boundary to extract spectral coordinates with explicit consideration of local object width and object part symmetry.
We have introduced a novel way to compute the adjacency weights between boundary nodes, based on the volume of intersection of their associated inscribed spheres. The use of this medially driven spectral coordinate leads to improved correspondences when compared against the use of conformal methods. When applied to 3D part segmentation the use of the medial surface to group together points on the boundary that are not necessarily close to one another in geodesic terms, but are tied to each other geometrically in terms of local object symmetry, has advantages. Finally, our object classification experiments using the medial spectral coordinates in addition to the raw data demonstrate that a vanilla deep neural network model can benefit from the incorporation of these additional spectral features.

In future work, we aim to examine whether it is possible to compute medial spectral coordinates implicitly, via a deep neural network model. We also hope to examine the potential of medial spectral coordinates for graph neural network models.

\bibliographystyle{unsrt}  
\bibliography{references}

\end{document}